%% file: main.tex
\documentclass[10pt]{article} % For LaTeX2e
\usepackage[preprint]{tmlr}

\input{math_commands.tex}

\usepackage{hyperref}
\usepackage{url}
\usepackage{longtable}
\usepackage{tikz}
\usepackage{pifont}
\usepackage{multirow}
\usepackage{booktabs}
\usepackage{tcolorbox}
\usepackage{enumitem}
\usepackage{array}
\usepackage{relsize}
\usepackage{inconsolata}
\usepackage{wrapfig}
\usepackage{siunitx}
\usepackage[dvipsnames]{xcolor}
\usepackage[table]{xcolor}

\tcbuselibrary{breakable}

\newcolumntype{C}[1]{>{\centering\arraybackslash}p{#1}}
\newcolumntype{R}[1]{>{\raggedleft\arraybackslash}p{#1}}
\newcolumntype{L}[1]{>{\raggedright\arraybackslash}p{#1}}

\definecolor{deepgreen}{rgb}{0.0, 0.5, 0.0}
\definecolor{ForestGreen}{rgb}{0.13, 0.55, 0.13}
\definecolor{LightPurple}{RGB}{170,143,210}
\definecolor{SteelBlue}{RGB}{70,130,180}
\newcommand{\xmark}{\textcolor{red}{\textbf{\ding{55}}}}
\newcommand{\cmark}{\textcolor{deepgreen}{\textbf{\ding{51}}}}
\newcommand{\customtexttt}[3]{{\fontsize{#1}{#2}\selectfont\texttt{#3}}}

\newcommand{\ctx}[1]{\cmark\rlap{\textsuperscript{\textsc{#1}}}}

\newcommand{\squishlist}{
\begin{list}{\footnotesize$\bullet$}
{\setlength{\itemsep}{0pt}
 \setlength{\parsep}{3pt}
 \setlength{\topsep}{3pt}       
 \setlength{\partopsep}{0pt}
 \setlength{\leftmargin}{1.5em}   
 \setlength{\labelwidth}{1em}
 \setlength{\labelsep}{0.5em}}
}
\newcommand{\squishend}{\end{list}}

\title{\textsf{SciTrek}: Evaluating and Improving Long-Context Numerical Reasoning over Scientific Articles}
\author{\name Miao Li, Alexander Gurung, Irina Saparina, Mirella Lapata \\ 
      \addr School of Informatics \\ University of Edinburgh \\
      \email miao.li@ed.ac.uk, alex.gurung@ed.ac.uk, i.saparina@sms.ed.ac.uk, mlap@inf.ed.ac.uk}

\def\month{MM}  % Insert correct month for camera-ready version
\def\year{YYYY} % Insert correct year for camera-ready version
\def\openreview{\url{https://openreview.net/forum?id=XXXX}} % Insert correct link to OpenReview for camera-ready version

\begin{document}

\maketitle

\begin{abstract}
We introduce \textsf{SciTrek}, a synthetic question-answering dataset for assessing and improving long-context numerical reasoning in large language models (LLMs). 
Existing long-context datasets with inputs beyond 64K tokens either target simple information retrieval or, when they do involve reasoning, rely on artificial contexts, while numerical reasoning remains largely overlooked in both cases.
% \alex{why do we need synthetic reasoning over artificial contexts? maybe don't need that phrase}
\textsf{SciTrek} addresses these limitations with questions that require numerical operations (e.g., counting, sorting, aggregation, and comparison) over collections of full-text scientific articles.
Questions are generated automatically by formulating them as SQL queries over a database of article metadata (titles, authors, and references), and ground-truth answers are obtained by executing these queries.
The underlying SQL provides a transparent, verifiable specification of the reasoning each question requires, enabling fine-grained error analysis. Furthermore, the proposed fully automated pipeline scales to arbitrary context lengths and dataset sizes with minimal human supervision, mitigating data contamination and supplying abundant data for post-training.
Extensive experiments show that frontier open-weight and proprietary LLMs struggle even with ostensibly simple questions: the best-performing model achieves only 46.5\% exact match at 128K tokens, and performance degrades steadily as contexts grow.
Fine-grained analysis reveals systematic weaknesses on citation-related questions and on compound logical conditions, particularly those involving negation.
Finally, post-training open-weight models on \textsf{SciTrek} improves their numerical reasoning in ways that generalise to out-of-domain long-context tasks.\footnote{Our dataset and code are available at \url{https://github.com/oaimli/SciTrek}.}
\end{abstract}

\section{Introduction}

Large language models (LLMs) now routinely support context windows of 128K tokens, with some exceeding 1M tokens \citep{gemini25_2025, gpt41_2025,qwen25_1m_2025}.\footnote{In this paper, K denotes 1,024 tokens and M denotes
1,024K.} However, models that accommodate such inputs do not necessarily
reason over them reliably, struggling, for instance, to integrate information
scattered across multiple documents~\citep{longbench_v2_2025, loong_2024}.
Diagnosing and addressing these failures requires data: progress in
long-context modelling has consistently depended on large, high-quality
datasets for both evaluation and training~\citep{long_context_survey_2025}.
While a growing number of such datasets have been
developed~\citep{babilong_2024, holistic_reasoning_2025, longbench_v2_2025,
helmet_2025}, existing efforts suffer from three key limitations.

First, although the inputs are sufficiently long, most existing datasets emphasize simple information retrieval rather than multi-step reasoning. Needle-In-A-Haystack tasks~\citep{needleinhaystack_2023, ruler_2024}, for instance, have been largely saturated by state-of-the-art long-context language models~\citep{gemini25_2025, gpt41_2025, qwen25_1m_2025}. 
Second, datasets that target  reasoning face a tradeoff between naturalness and scalability. Some rely on artificial contexts created by injecting noise into short passages~\citep{babilong_2024} (e.g., HotpotQA; \citealt{hotpotqa_2018}) or by synthesizing inputs with LLMs~\citep{holistic_reasoning_2025, mathhay_2024}, which limits ecological validity. 
% \alex{idk what ecological validity means} 
Other datasets are  built on natural long contexts but depend on substantial human annotation, which makes scaling to  rapidly expanding context windows difficult and
increases the risk of data contamination and
obsolescence~\citep{openscholar_2024, leval_2024, loong_2024, longbench_v2_2025}. 
Third, numerical reasoning --- counting, sorting, aggregation, and comparison --- remains largely unexplored in long-context datasets, despite being fundamental to real-world tasks~\citep{qdmr_2020, monaco_2025}. Moreover, because existing datasets rarely make explicit which reasoning skills each question targets, it is difficult to attribute model failures to specific capabilities.

In this paper, we introduce \textsf{SciTrek}, a question-answering dataset over scientific articles that addresses all three limitations. Scientific article collections provide a particularly suitable testbed for long-context numerical reasoning: they contain rich relational structure through citations, authorship, and cross-references, offer naturally occurring inputs, and scale seamlessly to long contexts as {their} size grows. 
% \alex{should we mention that raw articles is a natural input format?}
We generate questions requiring numerical operations, including counting, sorting, aggregation, and comparison, over core elements of scientific articles: titles, authors, and references. 
By relying on domain-general, readily identifiable elements rather than scientific content, the resulting tasks minimize dependence on domain-specific understanding while directly targeting long-context numerical reasoning itself.
Because these operations correspond to standard database commands and operators (Table~\ref{tab:sql_clauses}), we formulate them as SQL queries over article metadata, enabling automatic derivation of ground-truth answers through query execution without manual annotation. We then exploit LLM-based SQL-to-text generation~\citep{ li-etal-2024-bird, hui2024qwen2} to convert queries into natural-language questions, and ensure that their meaning is preserved through round-trip verification (SQL $\to$ question $\to$ SQL).

\begin{table}[t!]
\centering
\scalebox{0.98}{
\small
\begin{tabular}{L{0.45\linewidth}@{~~~}p{6.0cm}p{1.8cm}}
\toprule
\multicolumn{1}{c}{\textbf{Question}} & \multicolumn{1}{c}{\textbf{SQL Query}}	 & \multicolumn{1}{c}{\textbf{Answer}} \\
\midrule
\relscale{0.96}{What is the highest number of authors that any single article has?} & \customtexttt{8}{9}{SELECT MAX(author\_count) FROM articles} & \centerline{\relscale{0.95}{10}}\\
\midrule
\relscale{0.96}{What is the word count of the titles of articles, sorted by the number of authors in ascending order?} & \customtexttt{8}{9}{SELECT title\_word\_count FROM articles ORDER BY author\_count ASC} & \relscale{0.95}{9, 17, 5, 6, 9, 12}\\
\midrule
\relscale{0.96}{How many references do articles with exactly two authors have?} & \customtexttt{8}{9}{SELECT reference\_count FROM articles WHERE author\_count = 2} & \centerline{\relscale{0.95}{16}}\\
\midrule
\relscale{0.96}{What is the total number of words in the titles of all articles that have exactly 60 references?} & \customtexttt{8}{9}{SELECT SUM(title\_word\_count) FROM articles WHERE reference\_count $=$ 60} & \centerline{\relscale{0.95}{13}}\\
\midrule
\relscale{1.0}{What are the names of the authors who are either the first or second author of an article, listed in descending order of their position?} & \customtexttt{8}{9}{SELECT author\_name FROM article\_author WHERE author\_position < 2 ORDER BY author\_position DESC} & N. Shazeer, A. Vaswani\\
\midrule
\relscale{0.96}{How many articles are cited by other articles but do not cite any other articles?} & \customtexttt{8}{9}{SELECT COUNT(*) FROM articles WHERE article\_id NOT IN (SELECT article\_id\_citing FROM citing\_cited) AND article\_id IN (SELECT article\_id\_cited FROM citing\_cited)} & \centerline{\relscale{0.95}{2}} \\
\bottomrule
\end{tabular}
}
\caption{The {\textsf{\textsf{SciTrek}}} dataset: example questions
  with corresponding SQL queries and answers.} 
\label{tab:sql_examples}
\vspace{-5pt}
\end{table}

As an example, consider the first question in Table~\ref{tab:sql_examples}:
\textsl{``What is the highest number of authors that any single article has?''} Although the question seems simple, answering it requires locating the author list of every article in the context, counting the authors per
article, comparing these counts, and identifying the maximum. The numerical operations in \textsf{SciTrek} underpin complex questions in expert-curated
benchmarks~\citep{qdmr_2020, monaco_2025} and recur in human-written questions in OpenScholar~\citep{openscholar_2024},
LongBench~v2~\citep{longbench_v2_2025}, and Loong~\citep{loong_2024}
(Appendix~\ref{appendix:expert_written_questions}).

Our questions {are grounded in} surface-level metadata rather than scientific content, and some may not resemble queries a human would naturally pose. Nevertheless, they probe prerequisite capabilities: a model that cannot reason
accurately about such easily identifiable elements is unlikely to succeed on deeper analytical tasks. Indeed, when models are given the metadata as
database tables instead of full-text articles, accuracy rises sharply (Section~\ref{sec:evaluation_results}), indicating that the primary challenge lies in extracting and tracking relevant information across long contexts rather than in the numerical operations themselves.
\textsf{SciTrek} is diagnostic in two complementary ways: the table-article gap isolates the cost of reasoning over unstructured long contexts, and the underlying executable SQL queries provide an interpretable reference for fine-grained error analysis.

While our automated pipeline can generate an unlimited number of question-answer pairs, \textsf{SciTrek} provides a curated test set of 2,121 and a training set of 19,543 instances over scientific articles, spanning context lengths of 64K, 128K, 512K, and 1M tokens. We use the test set to conduct an extensive, fine-grained evaluation of frontier LLMs with context windows exceeding 128K tokens. The training set, in turn, is used to post-train open-weight models, improving their numerical reasoning on long-context tasks in other domains, including
LongBench~v2~\citep{longbench_v2_2025} and Loong~\citep{loong_2024}.
%\footnote{Appendix~\ref{appendix:example_questions_labeled} contains expert-written example questions on scientific articles, which likewise call for these capabilities.} 
Our contributions are as follows: 

% \begin{itemize}[leftmargin=1em,itemsep=1pt]
\squishlist

\item We develop a long-context numerical reasoning dataset grounded in scientific articles. The underlying SQL queries for each question enable fine-grained diagnosis of model failures, and the automated data curation pipeline scales effortlessly to longer contexts and larger datasets with minimal human effort.

\item Extensive evaluation reveals that frontier LLMs struggle even with the ostensibly simple reasoning questions in \textsf{SciTrek}, and that
performance declines sharply as context length grows. Fine-grained analysis traces these failures to citation-related questions, sorting, and compound logical conditions, particularly those involving negation.

\item Beyond its diagnostic value, \textsf{SciTrek} serves as an effective post-training corpus: fine-tuning open-weight LLMs on its training set improves their numerical reasoning, with gains that generalise to
out-of-domain long-context tasks.

%\item Beyond its diagnostic value, \textsf{SciTrek} can be used to improve the numerical reasoning of long-context language models via post-training. Our experiments show that the numerical reasoning capabilities that the models learn from \textsf{SciTrek} generalise to other out-of-domain long-context tasks.
% \end{itemize}
\squishend

\begin{table*}[t]
\small
\centering
\setlength{\tabcolsep}{3pt}
\scalebox{0.97}{
\begin{tabular}{l@{\hspace{-.1cm}}c@{~~}c@{~~}c@{~~}rc@{~~}p{7.4cm}}  \toprule
\textbf{Dataset} & \textbf{N-Ctx} & \textbf{Numerical} &
\textbf{Scalable} & \textbf{Length} & \textbf{Use} &
\hspace*{1cm}\textbf{Example Question} \\
\midrule
NeedleBench & \xmark & \xmark & \cmark & 128K & E & {\relscale{0.95}{What legendary item is hidden on Emerald Island?}} \\
Ada-LEval & \xmark & \xmark & \cmark & 128K & E & {\relscale{0.95}{What is the correct order of the segments?}} \\
BABILong & \xmark & \xmark &  \cmark & 10M & E &  {\relscale{0.95}{Where is Mary?}} \\
HELMET & \xmark & \xmark & \xmark & 128K & E &  {\relscale{0.95}{Who set the fire in one tree hill?}} \\
LIFBENCH & \xmark & \xmark & \cmark & 128K & E &  {\relscale{0.95}{Retrieve the entry at position 8th in the list.}} \\
RULER & \xmark & \xmark & \cmark & 128K & E &  {\relscale{0.95}{Find all variables that are assigned the value 12345.}} \\
OpenScholar & \ctx{s/m} & \xmark & \xmark & 3K$^*$ & E &  {\relscale{0.95}{Compile a list of reviews [...], and identify the most promising [...]}} \\
LongBench v2 & \ctx{s/m} & \xmark & \xmark & 128K & E &  {\relscale{0.95}{How long have I been living in my current apartment in Shinjuku?}} \\
LongMemEval & \ctx{m} &  \xmark & \xmark & 2M & E &  {\relscale{0.95}{How many bikes do I currently own?}} \\
L-Eval & \ctx{s/m} & \xmark & \xmark  & 200K & E & {\relscale{0.95}{How do I know when I should apply for Medicare?}} \\
HoloBench & \xmark  & \cmark & \cmark & 64K & E & {\relscale{0.95}{What are the names of wines and their corresponding grape types?}} \\
MathHay & \xmark  & \cmark & \cmark  & 128K & E & {\relscale{0.93}{What is the total number of points scored by LeBron [...] combined?}} \\
Loong & \ctx{m} & \cmark & \xmark & 250K & E & {\relscale{0.93}{Which company has the highest non-current assets?}} \\
\midrule
LoongRL & \xmark  & \xmark & \cmark & 16K & T & {\relscale{0.93}{When did Louis Deniset's political party form?}} \\
LongRLVR & \ctx{s} & \xmark & \cmark & 64K & T & {\relscale{0.93}{What is the name of the Bastard's father, what honor does he receive [...] to England?}} \\
\midrule
\textsf{SciTrek} (ours) & \ctx{m} & \cmark & \cmark & 1M & E, T&  {\relscale{0.93}{How many articles are cited by other articles but do not cite any other articles?}} \\
\bottomrule
\end{tabular}
}
\caption{Representative datasets for evaluating and post-training long-context language models. \mbox{\textbf{N-Ctx}}: whether the dataset uses naturally occurring input contexts; for natural contexts, we further indicate whether answering the  questions requires a single ($^{\textsc{S}}$) or multiple ($^{\textsc{M}}$) documents. \textbf{Numerical}: whether the dataset requires numerical reasoning over long contexts; \textbf{Scalable}: whether the dataset can be extended with minimal human effort (e.g.,~to longer contexts or larger datasets); \textbf{Length}: maximum supported input length in tokens; $^*$: length is based on the retrieved text rather than the full articles which are not available. \textbf{Use}: whether the dataset is intended for training (T),   evaluation (E), or both.} 
\label{tab:benchmarks}
\vspace{-10pt}
\end{table*}

\section{Related Work}

\paragraph{Retrieval-focused Long-context Datasets}
The increasing interest in long-context language models has driven the development of numerous datasets, mostly targeting benchmarking. Initial efforts primarily evaluate whether models can accurately retrieve specific information in long contexts, typically through Needle-In-A-Haystack tasks such as NeedleBench~\citep{needlebench_2024}. Subsequent benchmarks extend this paradigm by adding distractor information or scattering relevant content across the input, including AdaLEval~\citep{ada_leval_2024}, HELMET~\citep{helmet_2025}, LIFBENCH~\citep{lifbench_2025},
LongBench~\citep{longbench_2024}, and RULER~\citep{ruler_2024}. While these
datasets usefully stress-test information retrieval, they do not assess whether models can reason over the retrieved information.

\paragraph{Reasoning over Synthetic Contexts} A second line of work targets reasoning over long contexts but relies on
artificial inputs. BABILong~\citep{babilong_2024} embeds short reasoning tasks in long streams of unrelated text, HoloBench~\citep{holistic_reasoning_2025} considers numerical reasoning over various types of tables, and
MathHay~\citep{mathhay_2024} focuses on mathematical reasoning within long documents. However, they all construct their contexts synthetically --- by injecting noise into short passages or generating inputs with LLMs ---
so models may exploit distributional regularities of synthetic text that do not transfer to naturally occurring documents.

\paragraph{Reasoning over Natural Contexts} Several datasets combine reasoning with natural contexts, including
LongBench~v2~\citep{longbench_v2_2025}, LongMemEval~\citep{longmemeval_2025}, and LEval~\citep{leval_2024}; Loong~\citep{loong_2024} and
OpenScholar~\citep{openscholar_2024} evaluate reasoning over scholarly documents in particular. However, these benchmarks rely on expert-curated
questions and answers, which caps them at evaluation scale; LongBench~v2, for instance, comprises 503 questions in total,  far short of the quantities required for post-training, and manual curation cannot keep pace
with rapidly expanding context windows, increasing the risk of data contamination and obsolescence. Moreover, because the reasoning required to reach a correct answer is typically not made explicit, fine-grained failure
analysis remains difficult.

\paragraph{Synthetic Data as Post-training Corpora}
Most long-context datasets grounded in natural inputs focus on evaluation~\citep{loong_2024, longmemeval_2025, longbench_v2_2025}. For
post-training, LoongRL~\citep{loongrl_2026}, LongRLVR~\citep{longrlvr_2026},
and QwenLong1.5~\citep{qwenlong15_2025} develop datasets to improve long-context reasoning, but all rely on synthetic contexts, and none explicitly targets numerical reasoning over long inputs (see the discussion in Appendix~\ref{appendix:expert_written_questions}). Numerical
reasoning itself has been studied extensively in short contexts, most prominently in DROP~\citep{drop_2019}; \textsf{SciTrek} extends this line
of work to inputs several orders of magnitude longer. Together, this leaves a scarcity of public training data for numerical reasoning over
long, natural contexts.

Unlike existing datasets, \textsf{SciTrek} is, to our knowledge, the first long-context reasoning dataset that is simultaneously grounded in naturally occurring documents and automatically scalable. Its contexts are
full-text scientific articles, and its numerical-reasoning questions are generated automatically from article metadata (titles, authors, and references) by repurposing resources and models from text-to-SQL research~\citep{yu-etal-2018-spider, li-etal-2024-bird, hui2024qwen2} for SQL-to-question generation. Loong is  closest to \textsf{SciTrek} in that it targets numerical reasoning  over natural legal, financial, and scientific documents. However, its reliance on substantial manual annotation prevents automatic scaling, and its evaluation is limited to contexts  of up to~250K tokens. A detailed comparison between \textsf{SciTrek} and existing datasets is presented in Table~\ref{tab:benchmarks}.

\section{The \textsf{SciTrek} Dataset}
\label{sec:benchmark_construction}

\begin{figure*}[t!]
\centering
\includegraphics[width=1.0\textwidth]{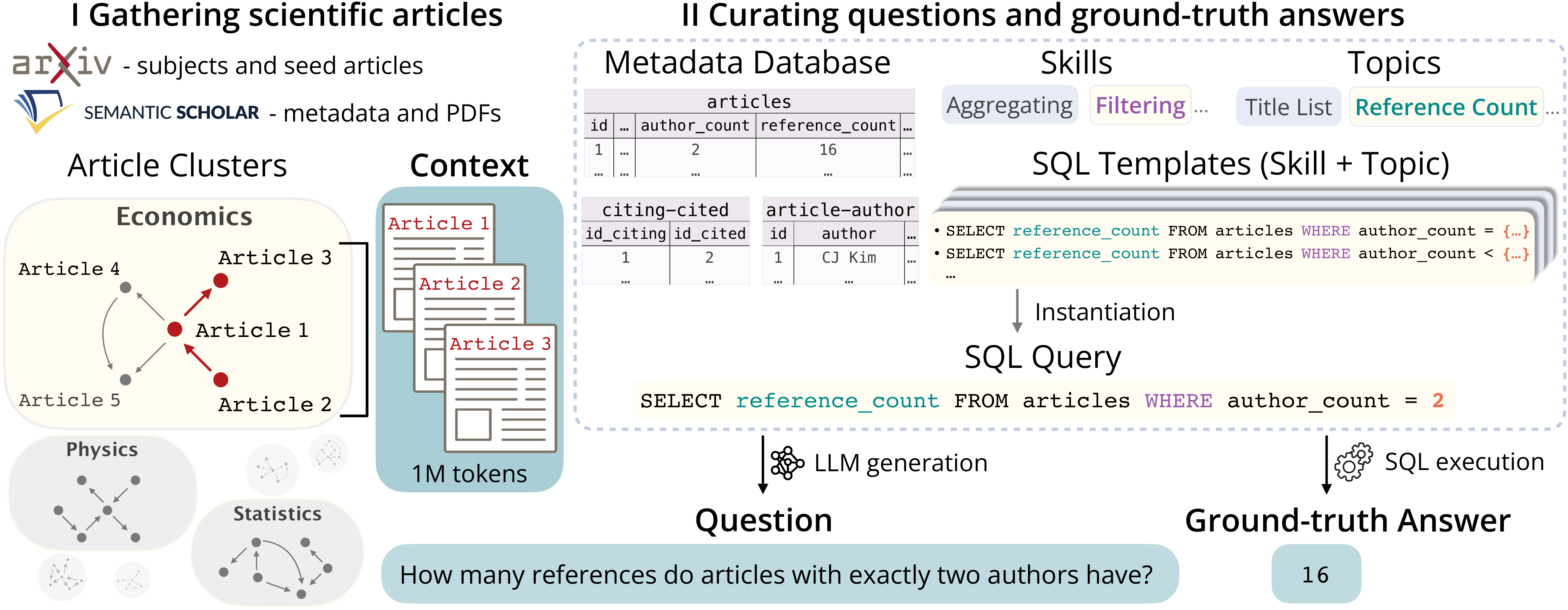}
\caption{Overview of \textsf{SciTrek} construction pipeline. We first gather scientific article collections of varying scales; then generate SQL queries and their answers from databases storing collection-specific metadata; and finally convert the SQL queries to natural language questions. The final dataset comprises three components highlighted in blue: full-text scientific articles as the input context, the natural language question, and the corresponding answer.} 
\label{fig:data_collection}
\vspace{-10pt}
\end{figure*}

Figure~\ref{fig:data_collection} illustrates the \textsf{SciTrek} curation pipeline:  we first assemble collections of scientific articles as contexts at varying lengths (i.e., 64K, 128K, 512K, and 1M), then construct databases representing article metadata and generate SQL queries with answers obtained via database execution. Finally, we convert the SQL queries into natural language questions. We describe each stage in detail below.

\subsection{Gathering Scientific Article Collections}
\label{sec:collections}

\textsf{SciTrek} is constructed from scientific articles sourced from Semantic Scholar.\footnote{\url{https://www.semanticscholar.org/}}  To ensure the representation of diverse topics, we obtain an initial set of seed articles from eight subjects: Computer Science (CS), Economics, Electronic Engineering (EE), Math, Physics, Biology, Finance, and Statistics.\footnote{Following the typology of subjects from \url{https://arxiv.org/}.} For each subject, we select two seed articles with more than~100 citations since~2020 and for each article, we retrieve related articles via the Semantic Scholar API.\footnote{Semantic Scholar provides APIs to retrieve scientific articles together with metadata including titles, authors, a reference list with articles that the current article cites, and a citation list with articles that cite the current article. Throughout this paper,  a reference is a bibliographic entry listed in the reference section of a scientific article, whereas a citation is an in-text acknowledgement of another’s work within the article.} To ensure broad coverage, we include two-hop related articles identified via Semantic Scholar's citation graph based on the reference and citation list. For each seed article, we randomly sample ten first-hop neighbors, and for each of these we further sample five second-hop neighbors forming an article cluster.   Since full texts are required, we filter out articles without PDFs. This process yields 16~article clusters comprising 662 scientific articles with PDFs across eight subjects (see step~I in Figure~\ref{fig:data_collection}). Finally, we convert the collected PDFs to Markdown using Marker.\footnote{\url{https://github.com/datalab-to/marker}}

From these clusters, we create article collections at varying context lengths for our question-answering task. By concatenating the Markdown texts, we produce collections of 64K, 128K, 512K, and 1M tokens.\footnote{We can easily construct collections with arbitrary lengths exceeding 1M tokens.}  Each collection is initialised with a randomly selected article, and further articles are appended until the target length is reached. We employ two expansion strategies: (1)~random sampling from the clusters, and (2)~traversal of the citation graph within each cluster (using both depth-first and breadth-first search) to construct collections that preserve citation relations among articles. Random groupings expose models to heterogeneous, cross-domain contexts, while citation-based groupings introduce richer relational structure that demands more sophisticated long-context reasoning. Each collection contains at least four articles, and no two collections share more than half of their articles. From the 662 scientific articles across 16 clusters, we construct 2,612 collections spanning all four length levels: 2,027 via random sampling and 585 via graph traversal.

\subsection{Creating Databases and SQL Queries}
\label{sec:databases_queries}

Once the article collections are assembled, we construct a database for each (see step~II in Figure~\ref{fig:data_collection}), focusing on a core subset of elements shared across all articles: titles, authors, and references. Based on these key elements, each database comprises three tables: \textit{articles}, \textit{article-author}, and \textit{citing-cited}. The \textit{articles} table contains metadata such as the title, reference count, and title word count. The \textit{article-author} table captures author information, including names and their positions in the corresponding author list. The \textit{citing-cited} table records citation relations among articles. A detailed description of the database tables is provided in Appendix~\ref{appendix:database_description}. This information is obtained from Semantic Scholar or derived via simple preprocessing, such as splitting titles and counting words. Manual corrections were applied as needed to align with the full-text Markdown articles.

\begin{wraptable}[14]{r}{0.44\textwidth}
\small
\centering
% \scalebox{0.98}{
\begin{tabular}{l@{~~}p{5cm}}
\toprule
\multicolumn{2}{c}{\textbf{SQL Commands}} \\
\midrule
Aggregating & \texttt{MAX}, \texttt{MIN}, \texttt{SUM}, \texttt{AVG}, \texttt{COUNT}, \texttt{DISTINCT} \\
Sorting & \texttt{ORDER BY}, \texttt{ASC}, \texttt{DESC}, \texttt{GROUP BY} \\
Filtering & \texttt{WHERE} \\
\bottomrule
\multicolumn{2}{c}{~}\\
\toprule
\multicolumn{2}{c}{\textbf{SQL Operators}} \\
\midrule
Comparison & \texttt{$=$}, \texttt{$>$}, \texttt{$<$}, \texttt{$>=$}, \texttt{$<=$}, \texttt{$<>$}, \texttt{LIKE} \\
Arithmetic & \texttt{$+$}, \texttt{$-$}, \texttt{$*$}, \texttt{/}, \texttt{\%} \\
Logical & \texttt{AND}, \texttt{NOT}, \texttt{OR}, \texttt{BETWEEN}, \texttt{IN} \\
\bottomrule
\end{tabular}
% }
\caption{Core commands and operators in SQL.}
\label{tab:sql_clauses}
%\vspace{-10pt}
\end{wraptable}

\begin{table*}[t]
% \vspace{-10pt}
\small
\centering
%\begin{minipage}[b]{0.48\linewidth}
%\centering
% \scalebox{0.95}{
\begin{tabular}{p{3.3cm}cp{10cm}}
\toprule
\textbf{Skill} & \textbf{Count} & \textbf{Example Query Template} \\
\midrule
Aggregating & 20 & \texttt{SELECT MAX(author\_count) FROM articles} \\
Sorting & 27 & \texttt{SELECT title\_word\_count FROM articles ORDER BY author\_count ASC} \\
Filtering & 107 & \texttt{SELECT author\_name FROM article\_author WHERE
  author\_position $=$ \{author-position\}}\\
Filtering+ Aggregating & 107 & \texttt{SELECT SUM(title\_word\_count) FROM articles WHERE reference\_count $=$ \{reference-count\}}\\
Filtering+ Sorting & 106 & \texttt{SELECT author\_count FROM articles WHERE title\_word\_count \% 2 $=$ 1 ORDER BY title\_word\_count DESC}\\
Relational Filtering & 20 & \texttt{SELECT COUNT(*) FROM articles WHERE article\_id NOT IN (SELECT article\_id\_citing FROM citing\_cited) AND article\_id IN (SELECT article\_id\_cited FROM citing\_cited)}\\
\bottomrule
\end{tabular}
% }
\caption{SQL templates representing different information
  processing skills. \{author-position\}, \{reference-count\} are placeholders.}
\label{tab:sql_templates}
\vspace{-10pt}
\end{table*}

The core SQL\footnote{Structured Query Language (SQL) is a standardized language for managing and querying relational databases.} commands summarized in Table~\ref{tab:sql_clauses} form the foundation for building SQL queries, often combined with operators (Comparison, Arithmetic, Logical) to create more complex queries (since \texttt{WHERE} is used to filter data, it always works in conjunction with SQL operators). Aside from the  SQL commands in Table~\ref{tab:sql_clauses} (Aggregating, Sorting, and Filtering), we define composite commands based on their combinations (i.e.,~Filtering+Aggregating and Filtering+Sorting). Using these, we manually create SQL query templates targeting different topics related to key elements of scientific articles, including Author Count, Author List, Reference Count, Title List, and Title Word Count. This process is illustrated in Figure~\ref{fig:data_collection} (right panel) and example templates are shown in Table~\ref{tab:sql_templates}; Note that some templates include placeholders to be instantiated with specific values.  Finally, to capture authorship and citation relations, we introduce Relational Filtering and the topics of Author Relation and Citation Relation, which specifically target authorship and citation relations. 

We collectively refer to the SQL commands in Table~\ref{tab:sql_templates} as information processing skills, since they test different information processing capabilities. We have various templates per skill, designed to be applicable across collections. For each collection, we randomly select 10~templates and instantiate them with collection-specific values for all placeholders. We then execute these queries against the corresponding database to generate ground-truth answers. 

\subsection{Converting SQL Queries to Natural Language}
We use Qwen2.5-Coder-32B-Instruct~\citep{hui2024qwen2} to convert SQL queries into natural language questions. To ensure queries and questions are meaning-preserving, we validate each generated question by converting it back to SQL and verifying that both queries produce identical results when executed against the collection database. For each SQL-collection pair, we repeat this process up to 10~times; if no valid question is obtained, we discard the query for that collection. Using the prompts in Appendix~\ref{appendix:prompt_templates}, we successfully generate natural language questions for 82.9\% of SQL-collection pairs. Following this, we obtain 2,121 test questions for evaluating models against the four context lengths defined in Section~\ref{sec:collections}, and 19,543 instances for training (see post-training experiments in Section~\ref{sec:long_context_post_training}). We split the data at the QA-pair level, ensuring that the QA pairs in the training, validation, and test sets are distinct, although source articles and SQL templates may overlap across splits. Table~\ref{tab:data_statistics_test} presents descriptive statistics for the \textsf{SciTrek} test set. For each context length, our test partition covers all information processing skills and question topics described in Section~\ref{sec:databases_queries}. The distribution of SQL commands and operators is provided in Appendix~\ref{appendix:database_description}, Table~\ref{tab:sql_distribution}.

\subsection{Data Quality Validation}

\begin{table}[t]
\small
\centering
\begin{tabular}{r S[table-format=3.0] S[table-format=2.1] S[table-format=2.1] S[table-format=2.1]}
\toprule
\textbf{Context Length} & {\#\textbf{Samples}} & {\#\textbf{Context Articles}} & {\textbf{Question Length}} & {\textbf{Gold-answer Length}} \\
\midrule
64K  & 112 &  4.2 & 14.6 &  6.2 \\
128K & 728 &  5.7 & 16.9 &  7.9 \\
512K & 667 & 22.7 & 18.1 & 30.9 \\
1M   & 614 & 46.3 & 18.6 & 63.6 \\
\bottomrule
\end{tabular}
\caption{Descriptive statistics for the \textsf{SciTrek} test set.
\textbf{Context Length}: target context length in tokens;
\textbf{\#Samples}: number of question--answer pairs; \textbf{\#Context Articles}: mean article count per context; \textbf{Question Length} and \textbf{Gold-answer Length}: mean length for questions and ground-truth answers in words.}
\label{tab:data_statistics_test}
\vspace{-10pt}
\end{table}

To validate \textsf{SciTrek}'s quality, we crowdsourced human annotations through Prolific.\footnote{https://www.prolific.com} Annotators (English native speakers from the US or the UK) were asked to answer natural language questions from a random sample of 120 instances, with 20 instances on average representing each of the six skills listed in Table~\ref{tab:sql_templates}.\footnote{Aggregating: 21, Sorting: 20, Filtering: 19, Filtering+Aggregating: 20, Filtering+Sorting: 21, and Relational Filtering: 19.} Crowdworkers answered questions using database tables rather than full-text articles, as documents spanning 1M tokens are impractical for humans to review.   Each instance was annotated by three participants following the same instructions as those used for model testing with database tables as context (Figure~\ref{fig:prompt_generation_meta_info}, Appendix~\ref{appendix:models}). In each annotation session, crowdworkers were presented with three questions over different contexts. To ensure annotation quality, each session included a quality control question. On average, annotators spent about 5.5 minutes per session and were compensated above the UK living wage, at \pounds{}12 per hour.

\begin{wraptable}[15]{r}{0.45\textwidth}
\centering
\small
\scalebox{0.95}{
\begin{tabular}{@{~}lcc@{~}}
\toprule
\textbf{Skill} & \textbf{Agree (\%)}	 & \textbf{Align (\%)} \\
\midrule
Aggregating & 85.7 & 85.7\\
Sorting & 85.0 & 80.0\\
Filtering & 85.7 & 85.7\\
Filtering+Aggregating & 95.0 & 85.0\\
Filtering+Sorting & 89.5 & 89.5\\
Relational Filtering & 89.5 & 73.7\\ 
\midrule
All & 88.3 & 83.3 \\
\bottomrule
\end{tabular}
}
\caption{Inter-annotator agreement on 120 sampled answers. Answers obtained from SQL execution are considered aligned if they match those provided by two or more annotators.}
\label{tab:data_quality_validation}
\end{wraptable}

We measured inter-annotator agreement and the alignment between database-derived answers (from SQL query execution) and human responses using exact match. Annotators were considered in agreement if two or more provided the same answer. Similarly, database answers were considered aligned with human responses if they matched the responses of two or more annotators. 
Table~\ref{tab:data_quality_validation} reports average agreement and alignment across 120 instances, showing strong inter-annotator agreement and high consistency between database-derived and human answers.

Manual inspection of all 120 questions given to annotators revealed that 3 of the 19 Relational Filtering questions were ambiguous, explaining the gap between alignment and agreement for this category in Table~\ref{tab:data_quality_validation}. While Qwen2.5-Coder-32B-Instruct generally converts SQL queries to natural language accurately, the complexity of Relational Filtering queries can introduce ambiguity.

\begin{table*}[t!]
\setlength{\tabcolsep}{3pt}
\centering
\small
\begin{tabular}{@{}lrcrrrrrrrr@{}}
\toprule
 & & & \multicolumn{4}{c}{\textbf{Full-text Articles}} & \multicolumn{4}{c@{}}{\textbf{Database Tables}} \\
\cmidrule(lr){4-7} \cmidrule(l){8-11}
\textbf{Model} & \textbf{Ctx} & \textbf{Think} & \textbf{F-64} & \textbf{F-128} & \textbf{F-512} & \textbf{F-1024} & \textbf{D-64} & \textbf{D-128} & \textbf{D-512} & \textbf{D-1024} \\
\midrule
Qwen-2.5-Coder + SQL execution & --- & --- & 38.4 & 35.3 & 34.1 & 25.6 & 90.6 & 89.4 & 86.8 & 87.5 \\
\midrule
Qwen2.5-7B-Instruct-1M         & 1M   & \xmark & 4.5  & 2.8  & 0.3 & 0.0 & 20.5 & 14.3 & 5.0  & 2.0 \\
Qwen2.5-14B-Instruct-1M        & 1M   & \xmark & 8.3  & 6.5  & 1.6 & 0.1 & 33.3 & 27.2 & 11.0 & 5.9 \\
Qwen3-4B-Instruct-2507         & 256K & \xmark & 2.1  & 7.2  & --- & --- & 25.9 & 16.9 & 6.9  & 2.8 \\
\rowcolor{gray!10}
Qwen3-4B-Thinking-2507         & 256K & \cmark & 41.1 & 29.3 & --- & --- & 90.2 & 83.1 & 71.8 & 52.3 \\
Qwen3-30B-A3B-Instruct-2507    & 256K & \xmark & 5.4  & 3.2  & --- & --- & 29.8 & 21.2 & 6.0  & 4.2 \\
\rowcolor{gray!10}
Qwen3-30B-A3B-Thinking-2507    & 256K & \cmark & 53.3 & 42.0 & --- & --- & 92.3 & 86.2 & 73.5 & 61.1 \\
Gemma-3-27B-IT                 & 128K & \xmark & 6.2  & 3.4  & --- & --- & 31.8 & 25.0 & 11.3 & 6.1 \\
Llama-4-Scout-17Bx16E-Instruct & 10M  & \xmark & 5.4  & 2.8  & 1.3 & 1.1 & 28.6 & 19.0 & 7.4  & 4.0 \\
Llama-3.3-70B-Instruct         & 128K & \xmark & 8.3  & 3.5  & --- & --- & 47.0 & 36.2 & 14.6 & 8.1 \\
\rowcolor{gray!10}
DeepSeek-R1-Distill-Llama-70B  & 128K & \cmark & 22.0 & 6.0  & --- & --- & 83.3 & 74.2 & 56.8 & 42.2 \\
\midrule
GPT-4.1                        & 1M   & \xmark & 21.1 & 11.7 & 3.9 & 2.5 & 69.3 & 53.8 & 24.8 & 16.1 \\
\rowcolor{gray!10}
Gemini 2.5 Pro                 & 1M   & \cmark & 41.7 & 26.0 & $\star$ & $\star$ & 91.7 & 83.5 & 55.4 & 31.5 \\
\rowcolor{gray!10}
o4-mini                        & 195K & \cmark & 61.0 & 46.5 & --- & --- & 95.2 & 87.8 & 79.4 & 72.6 \\
\bottomrule
\end{tabular}
\caption{Exact match (\%) on \textsf{SciTrek} for the SQL execution-based
method (top), open-weight (middle), and proprietary (bottom) models, under
two context settings: full-text scientific articles and the corresponding
database tables. \textbf{Ctx} is the maximum input length
each model supports; \textbf{Think} (\cmark, shaded rows) marks models
prompted with chain-of-thought, while non-thinking models (\xmark) answer
directly. F-* denotes test instances using full-text article collections of up to 64K, 128K, 512K, and 1M tokens. D-* denotes test data instances using the database tables corresponding to F-64, 128, 512, and 1024. For any given article collection, the database representation is substantially shorter (mean ${\sim}$2K tokens), and can therefore be passed to all models regardless of their context limit. --- : the model cannot process the given context size;
$\star$: not evaluated due to prohibitive computational cost.}
\label{tab:testing_results_em}
\vspace{-10pt}
\end{table*}

\section{Frontier-Model Evaluation}
\label{sec:evaluation_results}

We first use \textsf{SciTrek} to evaluate both proprietary and open-weight models from various families that support contexts exceeding 128K and have shown strong performance on language understanding and mathematical reasoning benchmarks such as MMLU~\citep{mmlu_2021} and MATH~\citep{math_2021}. 

For proprietary models, we include o4-mini\footnote{\url{https://developers.openai.com/api/docs/models/o4-mini}} and Gemini 2.5 Pro~\citep{gemini25_2025}, which are widely regarded as the leading models for long-context reasoning.
For open-weight models, we select the largest model we can feasibly run with our resources from each model family.\footnote{Experiments were conducted on 4 NVIDIA HGX H200s.} For example, we use Llama-4-Scout-17Bx16E-Instruct instead of Llama-4-Maverick-17B-128E-Instruct~\citep{llama4_2025}, and the distilled variant of DeepSeek-R1~\citep{deepseekai2024deepseekv2strongeconomicalefficient}. Table~\ref{tab:testing_results_em} provides an overview of the models we consider, which vary in parameter scale and supported context length (most were released within the past year). 
For detailed model descriptions and settings, refer to Appendix~\ref{appendix:models}. 
We evaluate models in two context settings: (1)~using the full-text scientific articles within a given collection as context, and (2)~using only the corresponding database tables.  We employ chain-of-thought prompting for thinking models, allowing them to output reasoning steps before the answer, while prompting non-thinking models to output the answer without any intermediate reasoning steps (actual prompts for these models are included in Appendix~\ref{appendix:models}). We assess performance using average exact match and F1, as the expected outputs are factual items with minimal variation, such as specific numbers, author names, or article titles. For higher confidence, reported metrics are averages across three runs. 

\begin{figure*}[t]
\centering
\includegraphics[width=1.0\textwidth, trim=0 68 0 0, clip]{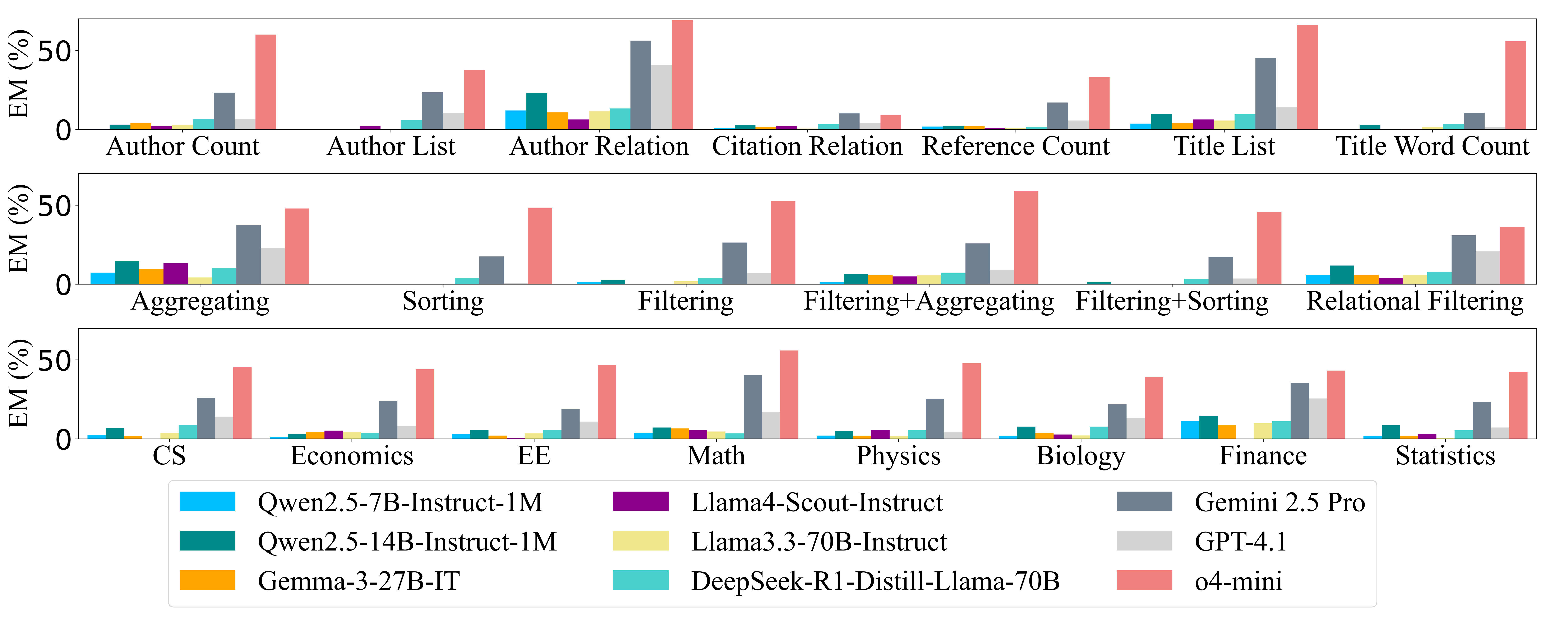}
%\vspace{-20pt}
\caption{Fine-grained analysis for best performing models in terms of exact match (EM). We examine how model performance varies across question topics (top), information processing skills (middle) and subjects (bottom) when using full-text scientific articles as context at an input length of 128K tokens.}
\label{fig:fine_grained_results_full}
\vspace{-5pt}
\end{figure*}

Model performance is summarised in Table~\ref{tab:testing_results_em}.
\textsf{SciTrek} proves challenging across all evaluated models,
especially with full-text articles as context: performance degrades for
every model as inputs grow longer and task size gets larger, and the same trend holds, more mildly,
with database tables. The single largest factor is thinking mode rather
than scale or provenance: at matched size, thinking variants outperform
their non-thinking counterparts by 30--50 EM points (e.g.,
Qwen3-30B-A3B: 5.4 vs.\ 53.3 at 64K), the best open-weight thinking model
surpasses Gemini~2.5~Pro, and o4-mini is strongest overall. Notably,
non-thinking models struggle even on the ${\sim}$2K-token database tables
(20.5--47.0 EM at 64K), indicating that a substantial part of the
difficulty lies in executing multi-step numerical operations, in addition
to locating and tracking information in long unstructured inputs.

\renewcommand{\arraystretch}{0.92}
\begin{wraptable}[20]{r}{0.62\textwidth}
\centering
\small
\begin{tabular}{@{}l@{}S@{~}S@{~}S@{~}S@{}}
\toprule
\textbf{Model} & {\textbf{\#Articles}} & {\textbf{Q-Len}} & {\textbf{SQL-Len}} & {\textbf{GA-Len}}\\
\midrule
Qwen2.5-7B-1M               & -.04 & \bfseries -0.15\rlap{$^*$} & 0.11 & -0.01 \\
Qwen2.5-14B-1M              & -.04 & \bfseries -0.16\rlap{$^*$} & 0.07 & -0.06 \\
Qwen3-4B-Instruct-2507      & -.01 & \bfseries -0.16\rlap{$^*$} & 0.05 & -0.03 \\
Qwen3-4B-Thinking-2507      & -.06 & -0.14\rlap{$^*$} & \bfseries -0.19\rlap{$^*$} & -0.01 \\
Qwen3-30B-A3B-Instruct-2507 & -.02 & \bfseries -0.15\rlap{$^*$} & 0.06 & 0.08 \\
Qwen3-30B-A3B-Think-2507    & -.12\rlap{$^*$} & -0.08 & \bfseries -0.19\rlap{$^*$} & -0.02 \\
Gemma-3-27B-IT              & -.08 & \bfseries -0.14\rlap{$^*$} & 0.02 & -0.08 \\
Llama-4-Scout               & -.10\rlap{$^*$} & \bfseries -0.17\rlap{$^*$} & -0.03 & -0.03 \\
Llama-3.3-70B               & -0.06 & \bfseries -0.16\rlap{$^*$} & -0.03 & -0.05 \\
DeepSeek-R1-Distill-Llama-70B & -0.15\rlap{$^*$} & \bfseries -0.16\rlap{$^*$} & -0.02 & -0.08 \\
\midrule
Gemini 2.5 Pro              & -0.16\rlap{$^*$} & \bfseries -0.22\rlap{$^*$} & -0.03 & -0.01 \\
GPT-4.1                     & -0.07 & \bfseries -0.21\rlap{$^*$} & 0.06 & 0.06 \\
o4-mini                     & -0.14\rlap{$^*$} & -0.09 & \bfseries -0.25\rlap{$^*$} & -0.06 \\
\bottomrule
\end{tabular}
\caption{Pearson correlation between per-instance exact match and four
properties of each instance (with full-text articles as context, 128K
tokens). \textbf{Q-Len:} question length, \textbf{SQL-Len:} SQL query length,  and \textbf{GA-Len:} gold answer length; all lengths are counted in space-separated words. $^*$: $p<0.05$.}
\label{tab:analysis_correlations}
\vspace{-10pt}
\end{wraptable}

We also evaluate an SQL execution-based baseline (first row of Table~\ref{tab:testing_results_em}): Qwen2.5-Coder-32B-Instruct converts each natural-language question into an SQL query, which is executed against the corresponding database. In the database-table setting, the query runs on the gold tables; in the full-text setting, we first construct the tables from titles, authors, and references extracted from the
articles by the same model.\footnote{Extraction prompts are given in
Appendix~\ref{appendix:prompt_templates}.} Execution on \emph{gold
tables} approaches ceiling ($\geq$86.8 EM) and constitutes an upper bound
for this extract-then-execute approach, whereas execution on \emph{extracted
tables} performs far worse (25.6--38.4 EM), showing that the main bottleneck in this pipeline is extracting accurate metadata from long unstructured text, rather than generating and executing the SQL query. Similar trends hold under F1
(Appendix~\ref{appendix:full_testing_results}).

We further conduct a fine-grained analysis for the best six models across question topics (e.g.,~author count, reference count, citation relations), skills (e.g.,~aggregating, sorting), and subjects (e.g.,~Economics, Biology) when using full-text articles as context (shown in Figure~\ref{fig:fine_grained_results_full}).  Model performance shows little variation by subject, which indicates that domain-specific content does not have an effect on our evaluation. However, most models struggle more with sorting tasks, while performing better on aggregation. Performance is lowest on citation-related questions (i.e.,~Citation Relation, Reference Count), and somewhat higher on author- and title-related questions. This is possibly because larger reasoning elements make it more challenging to get the answer. To understand which factors predict failure, Table~\ref{tab:analysis_correlations} reports correlations between per-instance exact match and four properties of each instance: the number of input articles, and the lengths (in words) of the question, the underlying SQL query, and the gold answer. All correlations are weak ($|r| \leq 0.25$), but a consistent pattern emerges. For non-thinking
models, performance correlates significantly only with question length,
suggesting sensitivity to surface complexity. For thinking models
(Qwen3-Thinking variants, o4-mini), the dominant correlate shifts to the
length of the underlying SQL query (a proxy for the structural
complexity of the required computation) while question length loses
significance for the strongest of them. Gold-answer length correlates with
performance for no model, indicating that failures are not driven simply
by questions demanding longer outputs.

Finally, to characterise \emph{how} models fail, we manually analysed 60
randomly selected instances at 128K input length on which all zero-shot
(and post-trained) models fail (10 per skill in
Table~\ref{tab:sql_templates}), and identified four recurring failure
patterns (further details in Appendix~\ref{sec:failure_patterns}): 
% \begin{enumerate}[wide=0pt, leftmargin=*, labelindent=0pt, itemsep=2pt, topsep=2pt]
    % \item 
(1) \textbf{Fallback to NULL.} Many open-weight models frequently output
``NULL'',  the answer our prompt reserves for cases where no answer can
be found in the context, particularly on filtering tasks, suggesting
they resort to this fallback rather than reasoning over the provided
input. 
% \item 
(2) \textbf{Format non-compliance.} Models often violate the specified answer format, especially on sorting questions, for instance
returning author lists when sorted author counts are requested, or lists
when scalar aggregates are expected; GPT-4.1 does so far more often than any other model. 
% \item 
(3) \textbf{Negation.} All models struggle with questions involving negation (e.g., ``not'', ``never'').
% \item 
(4) \textbf{Hallucination of specific facts.} As contexts grow, thinking
models (Qwen3-4B-Thinking-2507, Qwen3-30B-A3B-Thinking-2507,
DeepSeek-R1-Distill-Llama-70B) produce plausible high-level reasoning
plans yet hallucinate the specific facts needed to execute them, 
particularly counts, such as the number of reference items.
% \end{enumerate}

\begin{table}[t!]
\centering
\small
\setlength{\tabcolsep}{3pt}
\scalebox{.95}{
\begin{tabular}{lcSSSSSSS} 
\toprule
\textbf{Model}  & \textbf{Post-training} &  \textbf{Aggregating} & \textbf{Sorting} & \textbf{Filtering} & \textbf{F+A} & \textbf{F+S} & \textbf{Relational} & \textbf{Overall}  \\
\midrule
Qwen3-235B-A22B-Instruct & Zero-shot & 48.0 & 47.9 & 43.1 & 54.8 & 40.6 & 36.8 & 42.9 \\
\midrule
\multirow{3}{*}{Qwen2.5-7B-Instruct-1M} & Zero-shot & 7.8 & 0.0 & 1.2 & 2.6 & 0.0 & 6.1 & 3.3 \\
 & SFT & 14.7 & 0.0 & 4.2 & 13.4 & 2.4 & 30.4 & 15.4 \\
 & GRPO & 34.3 & 12.5 & 22.9 & 33.1 & 15.0 & 24.5 & 23.8 \\
\midrule
\multirow{3}{*}{Qwen3-4B-Instruct} & Zero-shot & 5.9 & 0.0 & 0.5 & 1.3 & 0.0 & 5.2 & 2.5 \\
 & SFT & 17.6 & 0.7 & 7.9 & 18.6 & 4.4 & 36.6 & 19.8 \\
% the results from ProxyCoT/simple_rl_openrlhf/scitrek_full_qwen3_4b_instruct_0527_simple_rl_test.jsonl on GAIL
 & GRPO & 43.1 & 21.5 & 31.5 & 47.2 & 25.8 & 30.4 & 32.9 \\
\midrule
\multirow{3}{*}{Gemma3-4B-IT} & Zero-shot & 3.9 & 0.0 & 0.0 & 1.7 & 0.0 & 1.3 & 1.0 \\
 & SFT & 3.9 & 0.0 & 3.5 & 12.6 & 3.1 & 21.8 & 11.6 \\
 & GRPO & 7.8 & 0.0 & 2.1 & 7.8 & 0.2 & 10.1 & 5.9 \\
\bottomrule
\end{tabular}
}
\caption{In-domain results of different numerical reasoning skills on the \textsf{SciTrek} test set (contexts$\leq$128K); models are evaluated using exact match (\%) in a zero-shot setting or after post-training on \textsf{SciTrek} with SFT or GRPO. We additionally report zero-shot results for Qwen3-235B-A22B-Instruct as a reference point at larger scale. \textbf{F+A}: Filtering + Aggregation; \textbf{F+S}: Filtering + Sorting; \textbf{Relational}: Relational Filtering.}
\label{tab:post_training_results_scitrek}
\vspace{-5pt}
\end{table}

\section{Post-training Experiments}
\label{sec:long_context_post_training}

Using the automated pipeline from Section~\ref{sec:benchmark_construction},
we generate a training set of 19,543 instances spanning the four context
lengths (64K, 128K, 512K, and 1M tokens). Because these instances exercise
diverse numerical operations over long, natural contexts without depending
on domain-specific knowledge, they are well suited to post-training
open-weight LLMs for long-context numerical reasoning. We evaluate both
in-domain gains on \textsf{SciTrek} and transfer to out-of-domain
long-context benchmarks.

We experiment with two well-established post-training techniques:
supervised fine-tuning (SFT) and reinforcement learning with verifiable
rewards (RLVR).\footnote{SFT trains models to map the input directly to the answer, whereas RLVR encourages models to generate reasoning traces before answering.} Owing to the high computational cost, we train on contexts of up
to 128K tokens (7,703 instances) and report results for three models from
different families: Qwen2.5-7B-Instruct-1M, Qwen3-4B-Instruct, and
Gemma3-4B-IT.\footnote{Due to computational resource constraints, our post-training experiments are confined to models under 10 billion parameters. However, since our data curation pipeline is agnostic to model architecture and scale, we expect the benefits to generalise to larger models as well.} For SFT, we train for 3~epochs with a batch size of~32, a
learning rate of $2\times10^{-6}$, and a warm-up ratio of~0.05. For RL, we
use GRPO~\citep{grpo_2024} with a summed EM and F1 reward, which we found
to improve accuracy over time while still providing signal on difficult
questions; GRPO has proven effective on similar verifiable long-context
tasks~\citep{grpo_2024, mroueh2025reinforcementlearningverifiablerewards,
gurung2025learning}. Owing to the computational cost of GRPO optimisation, we restrict our training to a single epoch (approximately five days) with a maximum generation budget of~2,048 tokens. All results are averaged over three responses per question to reduce sampling variance.

\paragraph{In-domain Results}
We first investigate whether post-training on \textsf{SciTrek} improves performance on \textsf{SciTrek} itself, and which numerical reasoning skills are more difficult to acquire. Table~\ref{tab:post_training_results_scitrek} compares model performance after post-training on \textsf{SciTrek} with the zero-shot setting. 
Both SFT and GRPO substantially improve performance, with GRPO yielding larger performance gains, particularly for Qwen3-4B-Instruct (comparable to Qwen3-235B-A22B-Instruct on aggregating-related tasks), though the smaller models generally remain well below the performance of Qwen3-235B-A22B-Instruct.  Across different numerical reasoning skills, sorting-related tasks are the most difficult to learn, whereas aggregation-related tasks benefit the most from post-training. This suggests sorting demands a qualitatively different capability that the two standard post-training techniques fail to unlock.

\begin{table}[t!]
\centering
\small
\scalebox{.95}{
\begin{tabular}{lcccc} 
\toprule
\textbf{Model}  & \textbf{Post-training} &  \textbf{LongBench v2} & \textbf{Loong-Academic} & \textbf{Loong-Financial} \\
\midrule
Qwen3-235B-A22B-Instruct & Zero-shot & 50.8~~~~~~~~~ & 49.2~~~~~~~~~ & 53.3~~~~~~~~~ \\
\midrule
\multirow{3}{*}{Qwen2.5-7B-Instruct-1M} & Zero-shot & 33.3~~~~~~~~~ & 31.3~~~~~~~~~ & 38.5~~~~~~~~~ \\
 & SFT & 34.4 \color{teal}{($+$1.1)} & 39.2 \color{teal}{($+$7.9)} & 39.5 \color{teal}{($+$1.0)} \\
 & GRPO & 35.7 \color{teal}{($+$2.4)} & 33.5 \color{teal}{($+$2.2)} & 42.0 \color{teal}{($+$3.5)} \\
\midrule
\multirow{3}{*}{Qwen3-4B-Instruct} & Zero-shot & 35.4~~~~~~~~~~  & 24.9~~~~~~~~~~ & 37.8~~~~~~~~~ \\
 & SFT & 13.3 \color{gray}{($-$22.1)} & ~46.4 \color{teal}{($+$21.5)} & 38.8 \color{teal}{($+$1.0)} \\
 & GRPO &  45.1 \color{teal}{($+$9.7)} & 34.3 \color{teal}{($+$9.4)} & 40.1 \color{teal}{($+$2.3)} \\
\midrule
\multirow{3}{*}{Gemma3-4B-IT} & Zero-shot & 26.8~~~~~~~~~ & 3.6~~~~~~~~ & 25.9~~~~~~~~~ \\
 & SFT & ~~~1.0 \color{gray}{($-$25.8)} & 11.8 \color{teal}{($+$8.2)} & 18.0 \color{gray}{($-$7.9)} \\
 & GRPO & 23.0 \color{gray}{($-$3.8)} & ~4.5 \color{teal}{($+$0.9)} & 25.2 \color{gray}{($-$0.7)} \\
\bottomrule
\end{tabular}
}
\caption{Out-of-domain transfer results on LongBench~v2 and Loong for models evaluated zero-shot or after post-training on \textsf{SciTrek} with SFT or GRPO. Performance on LongBench~v2 is multiple-choice accuracy
(\%); performance on Loong is answer quality on a 1--100 scale judged by
GPT-5~mini; higher is better in both cases;  We follow the evaluation
protocols of \citet{longbench_v2_2025} and \citet{loong_2024}.
Parenthesised values give the change relative to the same model's
zero-shot performance (\textcolor{teal}{teal}: improvement;
\textcolor{gray}{gray}: degradation). Qwen3-235B-A22B-Instruct is
included as a zero-shot reference for scale.}
\label{tab:post_training_results_ood}
\vspace{-5pt}
\end{table}

\paragraph{Out-of-domain Transfer}
We next evaluate whet her these gains extend beyond \textsf{SciTrek}
altogether, using two long-context benchmarks from different domains that
nonetheless require the numerical reasoning skills \textsf{SciTrek}
targets: Loong~\citep{loong_2024} and
LongBench~v2~\citep{longbench_v2_2025}.
Loong is designed to evaluate long-context language models through extended multi-document question answering with free-form answers, in which every document in each test case must be considered to derive the final answer. It spans four task types (Spotlight Locating, Comparison, Clustering, and Chain of Reasoning) across three domains (Financial Reports, Academic Papers, and Legal Cases) with context lengths ranging from 10K to beyond 200K tokens. We evaluate on the English-language samples with context lengths up to 128K tokens, without any further adaptation, using 309 instances from Financial Reports and 119 from Academic Papers.\footnote{The data instances from the Legal Cases domain are omitted, as they are entirely in Chinese.}  Unlike Loong, LongBench v2 is designed to evaluate long-context language models on deep understanding and reasoning across multiple real-world tasks with document, dialogue and code data. It consists of 503 challenging multiple-choice questions, with contexts ranging from 8K to 2M words. We evaluate on samples with context lengths up to 128K tokens.
We employ single-choice question accuracy (\%) and LLM-based answer quality (scored on a 1–100 scale by GPT-5 mini as a judge) as the evaluation metrics, following the evaluation protocols of ~\citet{longbench_v2_2025} and \citet{loong_2024}, respectively. 

Results in Table~\ref{tab:post_training_results_ood} show that post-training on \textsf{SciTrek} generally improves out-of-domain
performance, but the two methods behave very differently. GRPO yields
consistent gains for the Qwen2.5-7B and Qwen3-4B models across all three benchmarks (up to~$+$9.7 for Qwen3-4B on LongBench v2), bringing a 4B model within reach of the much larger Qwen3-235B zero-shot reference on LongBench~v2. SFT is far less stable: while producing the largest single gain ($+$21.5 for Qwen3-4B on Loong--Academic), it severely degrades LongBench~v2 performance for Qwen3-4B  and harms Gemma3-4B on two of three benchmarks. We attribute this asymmetry to output format: \textsf{SciTrek} and Loong both require free-form answers, whereas LongBench~v2 is multiple-choice, and SFT's stronger imprinting of \textsf{SciTrek}'s terse answer format appears to interfere with the multiple-choice setting, while GRPO, which optimises rewards over the model's own outputs, transfers across both formats. Gains are also uneven across base models: Gemma3-4B, the weakest zero-shot model, benefits little from GRPO (within $\pm$4 points on all three benchmarks) and suffers the largest SFT degradation (down to 1.0 on LongBench~v2), suggesting that transfer from \textsf{SciTrek} depends on sufficient base-model capability.

\begin{table}[t!]
\centering
\small
\scalebox{.95}{
\begin{tabular}{lccccc} 
\toprule
\textbf{Post-training Corpus} & \textbf{\#Samples} &  \textbf{LongBench v2} & \textbf{Loong-Academic} & \textbf{Loong-Financial} & \textbf{SciTrek} \\
\midrule
LoongRL & 21,024 & 36.1 & 32.4 & 33.8 & 25.4 \\
LongRLVR & 46,000 & 36.8 & \textbf{36.2} & 34.2 & 28.5 \\
\textsf{SciTrek} & ~7,703 & \textbf{45.1} & 34.3 & \textbf{40.1} & \textbf{32.9} \\
\bottomrule
\end{tabular}
}
\caption{Qwen3-4B-Instruct
is post-trained with GRPO (one epoch) on each corpus and evaluated on
LongBench~v2 (multiple-choice accuracy, \%), Loong (answer quality on
a 1--100 scale judged by GPT-5~mini; English subsets, contexts
$\leq$128K), and SciTrek test set (exact match, \%; contexts
$\leq$128K); higher is better. \#Samples: training instances
per corpus.}
\label{tab:post_training_results_existing_corpora}
\vspace{-10pt}
\end{table}

\paragraph{Comparison with Existing Post-training Corpora}
Finally, we compare \textsf{SciTrek} against two recent post-training
corpora for long-context reasoning: LoongRL~\citep{loongrl_2026}, which
expands Wikipedia-grounded short-context QA into 16K-token inputs by
inserting distractor documents and multi-hop key-value chains that
models must trace to recover the question, and
LongRLVR~\citep{longrlvr_2026}, which pairs natural documents of
8K--64K tokens (books, arXiv papers, code) with questions and answers
synthesised by Qwen3-235B-A22B-Instruct.\footnote{QwenLong-L1.5~\citep{qwenlong15_2025}
is excluded because its dataset was not publicly available at the time
of writing.} For a controlled comparison, we post-train the same model
(Qwen3-4B-Instruct) with the same method (GRPO, one epoch, 2{,}048-token
generation budget) on each corpus. As shown in
Table~\ref{tab:post_training_results_existing_corpora}, training on
\textsf{SciTrek} gives the best result on LongBench~v2 and
Loong--Financial, and remains competitive on Loong--Academic (34.3 vs.
LongRLVR's 36.2), despite using roughly a third of the training samples.
As
expected, it also attains the highest in-domain \textsf{SciTrek} accuracy. We attribute this efficiency primarily to data design.
\textsf{SciTrek} grounds reasoning in naturally occurring documents and
requires verifiable numerical operations over consistently identifiable
elements, providing a dense, checkable reward signal. LoongRL's
artificially assembled contexts, by contrast, likely reward
puzzle-specific tracing skills that transfer less readily to natural
documents, while LongRLVR's model-synthesised questions may inherit the
generator's distributional biases.

\section{Conclusion}
 
We introduced \textsf{SciTrek}, a synthetic dataset for evaluating and improving long-context numerical reasoning in LLMs, whose questions and
ground-truth answers are generated automatically as SQL queries over
metadata of scientific articles (titles, authors, and references).
Although the questions require only the ability to locate and manipulate
easily identifiable elements, our evaluation of thirteen open-weight and
proprietary LLMs shows they pose a substantial challenge: the best model
reaches 46.5\% exact match at 128K tokens, performance degrades sharply
with context length, and visible step-by-step reasoning (thinking mode) matters far
more than scale or provenance. Fine-grained analysis exposes systematic
failure modes, most notably compound logical conditions involving
negation and, for thinking models, hallucination of the specific facts (particularly counts) needed to execute otherwise sound reasoning plans. 
Post-training on \textsf{SciTrek} yields substantial in-domain gains
and transfers to out-of-domain long-context benchmarks, especially
under reinforcement learning; notably, it outperforms existing
long-context post-training corpora while using roughly a third of the
samples.
Finally, the generation pipeline extends naturally beyond titles,
authors, and references to richer article elements (e.g., tables and
figures) and to other domains whose entities have relational
structure (e.g., patient records, financial reports). Because
difficulty scales naturally with the underlying metadata,
\textsf{SciTrek} can grow to probe increasingly sophisticated reasoning as
context windows lengthen.
 
\newpage
\bibliography{main}
\bibliographystyle{tmlr}

\clearpage
\appendix
\section{Numerical Reasoning in Datasets with Expert-Written Questions}
\label{appendix:expert_written_questions}

Our numerical reasoning questions are constructed around atomic SQL operations, specifically sorting, filtering, and aggregating,  and operators covering comparison, arithmetic and logical reasoning,  shown in Table~\ref{tab:sql_clauses}. These capabilities are fundamental building blocks of complex, realistic tasks. QDMR~\citep{qdmr_2020} provides a widely used formalism, in which complex questions can be decomposed into atomic sub-questions that require filtering, aggregating, comparing, sorting, logical and arithmetic operations.  More recently, MoNaCo~\citep{monaco_2025}  demonstrates through manual decomposition of naturally occurring human-written questions that real-world complex questions in multi-document understanding frequently demand aggregation and arithmetic skills.

Our manual analysis of human-written or human-audited benchmark questions in OpenScholar~\citep{openscholar_2024}, LongBench v2~\citep{longbench_v2_2025}, and Loong~\citep{loong_2024} reveals that these questions implicitly require the same numerical reasoning capabilities targeted by  \textsf{SciTrek}. Example questions are shown in Table~\ref{tab:expert_questions}. Notably, such capabilities arise even in expert-written financial questions from LongBench v2, underscoring their relevance well beyond scientific text.

\begin{table*}[ht]
\setlength{\tabcolsep}{4pt}
\renewcommand{\arraystretch}{1.15}
\centering
\small
\begin{tabular}{@{}l >{\raggedright\arraybackslash}p{6.4cm} >{\raggedright\arraybackslash}p{3.1cm} >{\raggedright\arraybackslash}p{3.4cm}@{}}
\toprule
\textbf{Index} & \textbf{Question} & \textbf{Operations} & \textbf{Why} \\
\midrule
Q1-OS & What are the latest works on finetuning an auto-regressive LM for dense passage retrieval, and how does their performance compare with bi-directional encoders? & sorting, comparing & Order works by recency; compare reported performance across two model classes. \\
\addlinespace
Q2-OS & Which downstream task can be solved by AlphaFold3 but cannot be performed by ESM-3? & filtering, logical & Filter tasks by capability under a conjunction with negation. \\
\addlinespace
Q3-OS & Compared to 2023, how has the percentage of finished-goods apparel factories from countries other than Vietnam, China, and Cambodia changed in 2024? & filtering, aggregating, arithmetic, comparing & Filter by excluded countries; aggregate counts; compute percentages; compare across years. \\
\midrule
Q4-LB & In the financial reports of Apple and Samsung for 2022 and 2023, which company has a higher percentage of revenue from phones, and in what range do the differences between the two companies over the two years fall? & filtering, aggregating, arithmetic, comparing & Filter by company, year, and category; aggregate revenue; compute ratios and differences; compare. \\
\midrule
Q5-LG & [We ask you to] review the provided papers and construct a citation chain [\dots] that accurately reflects the sequential citation order among these documents. & sorting, filtering & Filter to citing--cited pairs; order them into a sequential chain. \\
\bottomrule
\end{tabular}
\caption{Expert-written questions from OpenScholar (OS),
LongBench~v2 (LB), and Loong (LG), reproduced verbatim (our elisions
marked [\dots]), with the \textsf{SciTrek} operations
(Table~\ref{tab:sql_clauses}) each implicitly requires and a brief
justification. These illustrate that the operations \textsf{SciTrek} targets recur
in naturally occurring benchmark questions.}
\label{tab:expert_questions}
\end{table*}

\section{\textsf{SciTrek} Database Construction} 
\label{appendix:database_description}
We construct a database with three tables for each article collection, using metadata from Semantic Scholar along with basic preprocessing. A description of these tables is provided in Table~\ref{tab:database_description}.
Our test dataset covers all SQL commands and operators that are listed in Table~\ref{tab:sql_clauses}. The detailed distribution is shown in Table~\ref{tab:sql_distribution}.

\begin{table*}[ht]
\setlength{\tabcolsep}{4pt}
\centering
\small
\scalebox{.95}{
\begin{tabular}{llcp{8cm}}
\toprule
\multicolumn{1}{l}{\textbf{Database Table}} & \multicolumn{1}{l}{\textbf{Column Name}} & \multicolumn{1}{c}{\textbf{Data Type}} & \multicolumn{1}{c}{\textbf{Description}}\\
\midrule
\multirow{5}{*}{\textit{articles}} & article$\_$id & String & The unique identifier of the article \\
& article$\_$title & String & The title of the article\\
& title$\_$word$\_$count & Integer & The number of words in the article's title (using spaces to determine word boundaries)\\
& author$\_$count & Integer & The number of authors in the article\\
& reference$\_$count & Integer & The number of references that are cited in the article \\
\midrule
\multirow{5}{*}{\textit{article-author}} & relation$\_$id & String & The unique identifier of the article-author relations \\
& article$\_$id & String & The identifier of the associated article\\
& author$\_$name & String & The name of the author/s\\
& author$\_$position & Integer & The position of the author in the author list (starting from 0 for the first author)\\
\midrule
\multirow{5}{*}{\textit{citing-cited}} & relation$\_$id & String & The unique identifier of the citation relations between two articles \\
& article$\_$id$\_$citing & String & The identifier of the article that cites another article\\
& article$\_$id$\_$cited & String & The identifier of the article that is cited by another article \\
\bottomrule
\end{tabular}
}
\caption{Description of database tables used to curate SQL queries and answers.}
\label{tab:database_description}
\end{table*}

\begin{table}[t]
\setlength{\tabcolsep}{4pt}
\centering
\small
\scalebox{0.95}{
\begin{tabular}{lrr@{\hspace{2.5em}}lrr}
\toprule
\textbf{Command/Operator} & \textbf{Count}& \textbf{\%}~~ & \textbf{Command/Operator} & \textbf{Count} & \textbf{\%}~~ \\
\midrule
\texttt{SELECT} & 2,121 & 100.00 & \texttt{GROUP BY} & 175 & 8.25 \\
\texttt{WHERE} & 1,821 & 85.86 & $\leq$ & 175 & 8.25 \\
$=$ & 1,032 & 48.66 & \% & 164 & 7.73 \\
\texttt{IN} & 682 & 32.15 & $\geq$ & 154 & 7.26 \\
\texttt{OR} & 616 & 29.04 & \texttt{NOT} & 154 & 7.26 \\
\texttt{ORDER BY} & 591 & 27.86 & $\neq$ & 153 & 7.21 \\
$<$ & 476 & 22.44 & \texttt{AVG} & 135 & 6.36 \\
$>$ & 450 & 21.22 & \texttt{MIN} & 133 & 6.27 \\
\texttt{COUNT} & 370 & 17.44 & \texttt{BETWEEN} & 106 & 5.00 \\
\texttt{ASC} & 301 & 14.19 & \texttt{SUM} & 102 & 4.81 \\
\texttt{DESC} & 287 & 13.53 & $/$ & 100 & 4.71 \\
\texttt{DISTINCT} & 280 & 13.20 & \texttt{LIKE} & 50 & 2.36 \\
$*$ & 215 & 10.14 & $+$ & 48 & 2.26 \\
\texttt{MAX} & 213 & 10.04 & $-$ & 23 & 1.08 \\
\texttt{AND} & 195 & 9.19 & & & \\
\bottomrule
\end{tabular}
}
\caption{Distribution of SQL commands and operators in our test data. Percentages reflect the prevalence of each command or operator among SQL queries and need not add to 100\%.}
\label{tab:sql_distribution}
\end{table}

%\section{Prompt Templates Used in \textsf{SciTrek} Construction}
%\label{appendix:prompts_benchmark_construction}
We use LLMs to extract titles, authors, and references from article texts; the corresponding prompts are shown in Appendix~\ref{appendix:prompt_templates}, Figure~\ref{fig:extraction_prompts}.
%For references specifically, we use two separate prompts: one to count references and one to detect the citation relationships between any two scientific articles, rather than extracting and outputting the full reference list directly.  Both operations are applied to article content in chunks of 16K tokens, as the extraction model (Qwen2.5-Coder-32B-Instruct in our experiments) does not reliably handle longer contexts. 

We generate natural language questions by prompting Qwen2.5-Coder-32B to convert SQL queries into natural language; the corresponding prompt is shown in Figure~\ref{fig:prompt_sql_to_nl}. The prompt for the reverse conversion, from natural language questions back to SQL queries, is shown in Figure~\ref{fig:prompt_nl_to_sql}.

\section{Experimental Settings and Evaluation}
\label{appendix:models}

Configuration details for all models are presented in Table~\ref{tab:models}, with default inference settings from their Huggingface repositories used throughout. 

Our evaluation covers both thinking and non-thinking models. For non-thinking models, we use two prompts: one to assess LLM capabilities against full-text articles (Figure~\ref{fig:prompt_generation_full_text}) and one for using the corresponding database tables as context (Figure~\ref{fig:prompt_generation_meta_info}).

The thinking models are Qwen3-4B-Thinking-2507, Qwen3-30B-A3B-Thinking-2507, DeepSeek-R1-Distill-Llama-70B, Gemini 2.5 Pro, and o4-mini, which differ in their thinking configurations: o4-mini uses the default \textit{medium} thinking effort with non-thinking prompts (Figure~\ref{fig:prompt_generation_full_text} and Figure~\ref{fig:prompt_generation_meta_info}), Gemini 2.5 Pro is allocated a budget of 512 tokens with non-thinking prompts (Figure~\ref{fig:prompt_generation_full_text} and Figure~\ref{fig:prompt_generation_meta_info}), and the three open-source models were prompted with the thinking prompt in Figure~\ref{fig:prompt_generation_reasoning} which allows models to output visible reasoning steps before the answer.

Finally, the prompt used by GRPO to generate reasoning traces is given in Figure~\ref{fig:prompt_generation_reasoning}.

\begin{table*}[t!]
\small
\centering
\begin{tabular}{@{}l r c r c@{}}
\toprule
\textbf{Model} & \textbf{\#Params} & \textbf{Think} & \textbf{Context Size} & \textbf{Release Date} \\
\midrule
\multicolumn{5}{@{}l}{\emph{Open-weight}} \\
Qwen2.5-7B-Instruct-1M         & 7B          & \xmark & 1,010,000  & Jan 2025 \\
Qwen2.5-14B-Instruct-1M        & 14B         & \xmark & 1,010,000  & Jan 2025 \\
Qwen3-4B-Instruct-2507         & 4B          & \xmark & 262,144    & Jul 2025 \\
\rowcolor{gray!10}
Qwen3-4B-Thinking-2507         & 4B          & \cmark & 262,144    & Jul 2025 \\
Qwen3-30B-A3B-Instruct-2507    & 30B (3B)    & \xmark & 262,144    & Jul 2025 \\
\rowcolor{gray!10}
Qwen3-30B-A3B-Thinking-2507    & 30B (3B)    & \cmark & 262,144    & Jul 2025 \\
Gemma-3-27B-IT                 & 27B         & \xmark & 131,072    & Mar 2025 \\
Llama-4-Scout-17Bx16E-Instruct & 109B (17B)  & \xmark & 10,485,760 & Apr 2025 \\
Llama-3.3-70B-Instruct         & 70B         & \xmark & 131,072    & Dec 2024 \\
\rowcolor{gray!10}
DeepSeek-R1-Distill-Llama-70B  & 70B         & \cmark & 131,072    & Jan 2025 \\
\midrule
\multicolumn{5}{@{}l}{\emph{Proprietary}} \\
GPT-4.1                        & ---         & \xmark & 1,047,576  & Apr 2025 \\
\rowcolor{gray!10}
Gemini 2.5 Pro                 & ---         & \cmark & 1,048,576  & Jun 2025 \\
\rowcolor{gray!10}
o4-mini                        & ---         & \cmark & 200,000    & Apr 2025 \\
\bottomrule
\end{tabular}
\caption{LLMs evaluated in our experiments. \textbf{\#Params}: total
parameters, with active parameters per token in brackets for
mixture-of-experts models; parameter counts are not disclosed for
proprietary models. \textbf{Think} (\cmark, shaded rows): models prompted
with chain-of-thought (Section~\ref{sec:evaluation_results}).
\textbf{Context Size}: maximum number of input tokens supported.}
\label{tab:models}
\end{table*}

\begin{table*}[th!]
\setlength{\tabcolsep}{3pt}
\centering
\small
\begin{tabular}{@{}lrcrrrrrrrr@{}}
\toprule
 & & & \multicolumn{4}{c}{\textbf{Full-text Articles}} & \multicolumn{4}{c@{}}{\textbf{Database Tables}} \\
\cmidrule(lr){4-7} \cmidrule(l){8-11}
\textbf{Model} & \textbf{Ctx} & \textbf{Think} & \textbf{F-64} & \textbf{F-128} & \textbf{F-512} & \textbf{F-1024} & \textbf{D-64} & \textbf{D-128} & \textbf{D-512} & \textbf{D-1024} \\
\midrule
Qwen-2.5-Coder + SQL execution & --- & --- & 51.5 & 49.9 & 50.8 & 52.7 & 94.0 & 91.8 & 89.4 & 91.6 \\
\midrule
Qwen2.5-7B-Instruct-1M         & 1M   & \xmark &  8.7 &  7.0 &  1.1 &  0.0 & 39.2 & 39.3 & 28.7 & 21.2 \\
Qwen2.5-14B-Instruct-1M        & 1M   & \xmark & 20.6 & 17.5 &  7.4 &  1.0 & 58.2 & 53.9 & 44.8 & 34.8 \\
Qwen3-4B-Instruct-2507         & 256K & \xmark &  7.2 &  7.3 & ---  & ---  & 47.2 & 46.0 & 38.9 & 31.1 \\
\rowcolor{gray!10}
Qwen3-4B-Thinking-2507         & 256K & \cmark & 50.2 & 42.3 & ---  & ---  & 92.7 & 85.8 & 79.4 & 63.6 \\
Qwen3-30B-A3B-Instruct-2507    & 256K & \xmark & 15.0 & 16.4 & ---  & ---  & 55.6 & 53.8 & 44.7 & 42.0 \\
\rowcolor{gray!10}
Qwen3-30B-A3B-Thinking-2507    & 256K & \cmark & 62.5 & 55.1 & ---  & ---  & 94.4 & 88.3 & 80.8 & 74.5 \\
Gemma-3-27B-IT                 & 128K & \xmark & 22.6 & 14.7 & ---  & ---  & 52.8 & 52.5 & 47.8 & 40.7 \\
Llama-4-Scout-17Bx16E-Instruct & 10M  & \xmark & 20.3 & 17.4 & 15.6 & 14.8 & 49.8 & 44.9 & 40.5 & 37.4 \\
Llama-3.3-70B-Instruct         & 128K & \xmark & 25.3 & 15.1 & ---  & ---  & 65.8 & 63.5 & 53.7 & 49.5 \\
\rowcolor{gray!10}
DeepSeek-R1-Distill-Llama-70B  & 128K & \cmark & 29.0 &  9.0 & ---  & ---  & 85.5 & 77.6 & 68.0 & 62.3 \\
\midrule
GPT-4.1                        & 1M   & \xmark & 36.0 & 29.7 & 22.3 & 19.6 & 82.8 & 73.3 & 61.8 & 56.3 \\
\rowcolor{gray!10}
Gemini 2.5 Pro                 & 1M   & \cmark & 58.1 & 48.8 & $\star$ & $\star$ & 95.3 & 88.3 & 79.0 & 69.8 \\
\rowcolor{gray!10}
o4-mini                        & 195K & \cmark & 74.0 & 63.8 & ---  & ---  & 97.2 & 89.2 & 87.0 & 86.5 \\
\bottomrule
\end{tabular}
\caption{F1 (\%) on \textsf{SciTrek} for the SQL execution-based
method (top), open-weight (middle), and proprietary (bottom) models, under
two context settings: full-text scientific articles and the corresponding
database tables. \textbf{Ctx} is the maximum input length
each model supports; \textbf{Think} (\cmark, shaded rows) marks models
prompted with chain-of-thought, while non-thinking models (\xmark) answer
directly. F-* denotes test instances using full-text article collections of up to 64K, 128K, 512K, and 1M tokens. D-* denotes test data instances using the database tables corresponding to F-64, 128, 512, and 1024. For any given article collection, the database representation is substantially shorter (mean ${\sim}$2K tokens), and can therefore be passed to all models regardless of their context limit. --- : the model cannot process the given context size;
$\star$: not evaluated due to prohibitive computational cost. Exact match
results are in Table~\ref{tab:testing_results_em}.}
\label{tab:testing_results_f1}
\end{table*}

%\section{Prompt Templates Used in Extraction of Titles, Authors, and Reference Items}
%\label{appendix:prompts_element_extraction}
%We use LLMs to extract titles, authors, and references from article texts; the corresponding prompts are shown in Table~\ref{tab:extraction_prompts}.
%For references specifically, we use two separate prompts: one to count references and one to detect the citation relationships between any two scientific articles, rather than extracting and outputting the full reference list directly.  Both operations are applied to article content in chunks of 16K tokens, as the extraction model (Qwen2.5-Coder-32B-Instruct in our experiments) does not reliably handle longer contexts. 

\section{Evaluation Results in Terms of F1}
\label{appendix:full_testing_results}

Table~\ref{tab:testing_results_f1} reports results under F1. The ranking
of models is unchanged from exact match
(Table~\ref{tab:testing_results_em}), confirming that our findings are not
an artefact of a strict matching criterion, but absolute scores are
substantially higher (e.g., o4-mini improves from 46.5 to 63.8 at 128K
tokens with full-text articles). The gap between the two metrics is
largest for questions with list-valued answers, as a sorted list with a single incorrect element receives a score of zero in terms of exact match, regardless of how close the answer is. This indicates that models frequently retrieve most of a required set or list while failing to
produce it exactly. 
Two patterns from the main results also
persist under F1: thinking models outperform their non-thinking
counterparts by a wide margin, and performance degrades with context
length for every model. The SQL execution-based method is the sole
exception: because it processes articles in 16K-token chunks, its
performance is essentially flat across collection sizes.

\section{Failure Patterns}
\label{sec:failure_patterns}

This appendix provides the per-skill breakdowns behind the four failure
patterns summarised in Section~\ref{sec:evaluation_results}. All figures
are computed over the 60 manually inspected instances at 128K input
length (10 per skill in Table~\ref{tab:sql_templates}) on which every
zero-shot and post-trained model fails, except for
Table~\ref{tab:performance_negation}, which reports exact match over all
132 negation instances in the test set.

\textbf{Fallback to NULL and format non-compliance.}
Table~\ref{tab:failure_breakdown} reports both patterns by skill. NULL
fallback is concentrated in the smaller open-weight models
(Qwen2.5-7B: 90\% on Aggregating, Qwen3-4B-Instruct: 90\% on
Filtering+Sorting) and is markedly rarer for thinking models and
proprietary systems; post-training removes it almost entirely (0\% in
every cell for GRPO). Format violations follow a different distribution:
they are worst for GPT-4.1 (30--50\% across Aggregating, Sorting, and
Filtering) and Gemma-3-27B, and, unlike NULL fallback, do not correlate
with model size.

\textbf{Negation.} Our test set contains 132 negation instances,
distributed across Filtering (11), Filtering+Aggregating (15),
Filtering+Sorting (12), and Relational Filtering (94).
Table~\ref{tab:performance_negation} shows exact match on this subset:
seven of thirteen models score below 5\% overall, and no model exceeds
19\%. The Relational Filtering column, which carries most instances,
is where scores are lowest for every model (${\leq}6.4$\%), indicating
that negation compounds with relational reasoning rather than being an
independent difficulty.

\textbf{Hallucination of specific facts.} Inspecting the reasoning traces
of Qwen3-4B-Thinking-2507, Qwen3-30B-A3B-Thinking-2507, and
DeepSeek-R1-Distill-Llama-70B, we find that the high-level plans are
typically correct (e.g.,~identify the relevant articles, count the target
items, compare the counts) while the intermediate quantities are not.
Counting reference items is the most common point of failure: models
state a specific reference count for an article and proceed to reason
correctly from that erroneous value.

\begin{table*}[t!]
\centering
\small
\begin{tabular}{@{}l rrrrrr rrrrrr@{}}
\toprule
 & \multicolumn{6}{c}{\textbf{Fallback to NULL}} & \multicolumn{6}{c@{}}{\textbf{Format Non-compliance}} \\
\cmidrule(lr){2-7} \cmidrule(l){8-13}
\textbf{Model} & \textbf{A} & \textbf{S} & \textbf{F} & \textbf{F+A} & \textbf{F+S} & \textbf{R}
               & \textbf{A} & \textbf{S} & \textbf{F} & \textbf{F+A} & \textbf{F+S} & \textbf{R} \\
\midrule
Qwen2.5-7B-Instruct-1M      & 90 & 10 & 80 & 10 & 40 & 70 &  0 &  0 &  0 &  0 &  0 & 10 \\
Qwen2.5-14B-Instruct-1M     &  0 &  0 & 30 & 30 & 30 & 80 & 10 & 30 &  0 & 20 & 30 & 10 \\
Qwen3-4B-Instruct-2507      & 50 & 60 & 80 & 10 & 90 & 90 &  0 &  0 &  0 &  0 &  0 &  0 \\
\rowcolor{gray!10}
Qwen3-4B-Thinking-2507      &  0 &  0 & 20 & 20 & 10 & 40 &  0 &  0 &  0 &  0 &  0 &  0 \\
Qwen3-30B-A3B-Inst.-2507    &  0 &  0 &  0 & 20 &  0 & 70 & 30 &  0 & 50 & 40 & 10 & 10 \\
\rowcolor{gray!10}
Qwen3-30B-A3B-Think.-2507   &  0 &  0 &  0 & 30 & 10 & 30 &  0 & 10 &  0 &  0 &  0 &  0 \\
Gemma-3-27B-IT              & 10 &  0 &  0 & 40 & 10 & 40 & 20 & 30 & 40 & 10 &  0 & 30 \\
Llama-4-Scout-Instruct      & 10 & 10 & 10 & 60 & 30 & 80 & 10 & 30 & 10 & 20 &  0 &  0 \\
Llama-3.3-70B-Instruct      & 20 & 30 & 40 & 70 & 30 & 70 &  0 & 10 & 10 & 10 & 10 & 10 \\
\rowcolor{gray!10}
DeepSeek-R1-Dist.-Llama-70B &  0 &  0 & 10 & 10 &  0 &  0 &  0 &  0 &  0 &  0 &  0 &  0 \\
GPT-4.1                     &  0 &  0 &  0 &  0 &  0 & 10 & 50 & 40 & 40 & 30 & 10 & 10 \\
\rowcolor{gray!10}
Gemini 2.5 Pro              &  0 &  0 &  0 & 20 &  0 & 30 & 10 &  0 &  0 & 10 &  0 &  0 \\
\rowcolor{gray!10}
o4-mini                     &  0 &  0 &  0 & 30 & 10 & 20 & 10 & 10 &  0 &  0 &  0 &  0 \\
\midrule
Qwen2.5-7B (SFT)            &  0 &  0 & 20 & 70 & 20 &  0 &  0 &  0 &  0 &  0 &  0 &  0 \\
Qwen2.5-7B (GRPO)           &  0 &  0 &  0 &  0 &  0 &  0 & 10 &  0 &  0 &  0 & 10 &  0 \\
\bottomrule
\end{tabular}
\caption{Proportion of inspected samples (\%) exhibiting the two
output-level failure patterns, by information-processing skill.
\textbf{A}: Aggregating; \textbf{S}: Sorting; \textbf{F}: Filtering; \textbf{R}: Relational Filtering.
Shaded rows are thinking models; the final block shows post-trained
variants of Qwen2.5-7B-Instruct-1M.}
\label{tab:failure_breakdown}
\end{table*}

\begin{table}[t!]
\centering
\small
% \scalebox{0.91}{
\begin{tabular}{lrrrrr}
\toprule
\multicolumn{1}{c}{\textbf{Model}}& \multicolumn{1}{c}{\textbf{F}} & \multicolumn{1}{c}{\textbf{F+A}} & \multicolumn{1}{c}{\textbf{F+S}} & \multicolumn{1}{c}{\textbf{R}} & \multicolumn{1}{c}{\textbf{All}}  \\
\midrule
Qwen2.5-7B-Instruct-1M & 0 & 0 & 0 & 0 & 0\\
Qwen2.5-14B-Instruct-1M & 0 & 20.0 & 0 & 0.2 & 3.8 \\
Qwen3-4B-Instruct-2507 & 0 & 0 & 0 & 0 & 0\\
Qwen3-4B-Thinking-2507 & 0 & 53.3 & 8.3 & 1.1 & 9.1\\
Qwen3-30B-A3B-Inst.-2507 & 0 & 0 & 0 & 0 & 0\\
Qwen3-30B-A3B-Thin.-2507 & 0 & 66.7 & 16.7 & 1.1 & 12.9\\
Gemma-3-27B-IT & 0 & 6.7 & 0 & 1.1 & 1.5\\
Llama-4-Scout-Instruct & 0 & 13.3 & 0 & 2.1 & 3.0\\
Llama-3.3-70B-Instruct & 0 & 6.7 & 0 & 0 & 0.8\\
DeepSeek-R1-Dist.-Llama-70B & 0 & 13.3 & 0 & 0 & 3.0\\
Gemini 2.5 Pro & 0 & 20.0 & 8.3 & 6.4 & 9.8\\
GPT-4.1  & 0 & 13.3 & 0 & 3.2 & 4.5\\
o4-mini  & 0 & 60.0 & 50.0 & 5.3 & 18.9 \\ \hline
Qwen2.5-7B-Inst.-1M (SFT) & 0 & 6.7 & 0 & 10.6 & 9.1\\
Qwen2.5-7B-Inst.-1M (GRPO) & 0 & 33.3 & 8.3 & 4.3 & 8.3 \\

\bottomrule
\end{tabular}
% }
\caption{Model performance on questions involving negation (e.g.,~``not" or ``never"), reported in terms of exact match (\%). The first block shows zero-shot models; the second block shows supervised versions of Qwen2.5-7B-Instruct-1M. \textbf{A} abbreviates Aggregation, \textbf{S} stands for Sorting, \textbf{F} for Filtering, and \textbf{R} for Relational Filtering.}
\label{tab:performance_negation}
\end{table}

\section{Prompt Templates}
\label{appendix:prompt_templates}

This appendix collects the prompt templates used throughout
\textsf{SciTrek}. Placeholders instantiated at run time are marked in
\textcolor{red}{red}. Figure~\ref{fig:extraction_prompts} contains the
four templates used to extract article metadata from full text: title
extraction, author extraction, reference counting, and detection of
citation relationships between article pairs. Each is applied to
16K-token chunks of an article, as the extraction model
(Qwen2.5-Coder-32B-Instruct) does not reliably handle longer inputs; for
references we count entries and detect citation links separately rather
than reproducing the full reference list, which is more robust than
extracting bibliographies verbatim.

Figures~\ref{fig:prompt_sql_to_nl} and~\ref{fig:prompt_nl_to_sql} contain
the templates for the two directions of query-question conversion. Both
present the same database schema, so that a question generated from an
SQL query in Figure~\ref{fig:prompt_sql_to_nl} can be converted back to
SQL under identical assumptions in Figure~\ref{fig:prompt_nl_to_sql};
this round-trip allows us to verify that generated questions
preserve the meaning of the original query
(Section~\ref{sec:benchmark_construction}). All prompts require JSON output, which we
enforce with constrained decoding.

Figures~\ref{fig:prompt_generation_full_text} and \ref{fig:prompt_generation_meta_info}
contain the templates used to evaluate models on \textsf{SciTrek}. The two differ only in how the context is presented: the first supplies the
full-text articles, the second the corresponding database tables, while
the answer-format instructions are identical, so that the two settings
differ in input representation alone.
Figure~\ref{fig:prompt_generation_reasoning} gives the variant used for
thinking models and for GRPO training, which additionally asks the model
to reason step by step and place its final answer in
\texttt{\textbackslash boxed\{\}}; it is otherwise identical to the
non-thinking prompt.

\begin{figure*}[th!]
\small
\begin{tcolorbox}[colback=gray!10!white, colframe=NavyBlue,
  title=Instruction for Title Extraction, fonttitle=\bfseries,
  halign title=flush center]
You are given a chunk of a scientific article.\\
\{\textcolor{red}{context}\}\\[2pt]
If the article title is included in the given text, please locate and extract the title.\\
You must output the title in the following JSON format, and do not include any extra text.\\
\{``title": ``the extracted title"\}\\
If there is no article title in the given text, just output \{``title": ``"\}
\end{tcolorbox}

\begin{tcolorbox}[colback=gray!10!white, colframe=NavyBlue,
  title=Instruction for Author Extraction, fonttitle=\bfseries,
  halign title=flush center]
You are given a chunk of a scientific article.\\
\{\textcolor{red}{context}\}\\[2pt]
If the author list of the article is included in the given text, please locate and extract the author list.\\
You must output the author list in the following JSON format, and do not include any extra text.\\
\{``authors": ``the author list separated with commas"\}\\
If there is no author list in the given text, just output \{``authors": ``"\}
\end{tcolorbox}

\begin{tcolorbox}[colback=gray!10!white, colframe=NavyBlue,
  title=Instruction for Counting References, fonttitle=\bfseries,
  halign title=flush center]
You are given a chunk of a scientific article.\\
\{\textcolor{red}{context}\}\\[2pt]
If a complete or partial reference section is included in the given text, please count the number of references.\\
You must output the reference count in the following JSON format, and do not include any extra text.\\
\{``reference\_count": ``the number of references"\}\\
If there is no reference section in the given text, just output \{``reference\_count": ``0"\}
\end{tcolorbox}

\begin{tcolorbox}[colback=gray!10!white, colframe=NavyBlue,
  title=Instruction for Detecting Citation Relationships, fonttitle=\bfseries,
  halign title=flush center]
You are given the title of one scientific article (Article~A) and a chunk of another scientific article (Article~B).\\[2pt]
The title of Article~A: \{\textcolor{red}{title}\}\\
The chunk of Article~B: \{\textcolor{red}{context}\}\\[2pt]
Please check whether Article~A is cited by Article~B; that is, check whether the title of Article~A appears in the given chunk of Article~B.\\
You must output your answer in the following JSON format, and do not include any extra text.\\
\{``a\_cited\_by\_b": ``true or false"\}
\end{tcolorbox}
\caption{Prompt templates for extracting titles, authors, and references
from article text. \{\textcolor{red}{context}\} and
\{\textcolor{red}{title}\} are placeholders for the input article chunk
and the article title.}
\label{fig:extraction_prompts}
\end{figure*}

\begin{figure*}[!ht]
\small
  \begin{tcolorbox}[colback=gray!10!white,colframe=NavyBlue,title=Instruction to Convert SQL Queries to Natural Language Questions,fonttitle=\bfseries, halign title=flush center]
    {You are given a database with three tables: \textit{articles}, \textit{article-author}, and \textit{citing-cited}.}\\
    
    {The \textit{articles} table contains the following columns:}
    \begin{itemize}[noitemsep]
    \item[-] article$\_$id (String): the unique identifier of the article;
    \item[-] article$\_$title (String): the title of the article;
    \item[-] title$\_$word$\_$count (Integer): the number of words in the article's title (using spaces to determine word boundaries);
    \item[-] author$\_$count (Integer): the number of authors for the article;
    \item[-] reference$\_$count (Integer): the number of references cited in the article.
   \end{itemize} 
   
    {The \textit{article-author} table contains the following columns:}
    \begin{itemize}[noitemsep]
    \item[-]relation$\_$id (String): the unique identifier of the article-author relationship;
    \item[-]article$\_$id (String): the identifier of the associated article;
    \item[-]author$\_$name (String): the name of the author;
    \item[-]author$\_$position (Integer): the position of the author in the author list (starting from 0 for the first author).
    \end{itemize}
    
    {The \textit{citing-cited} table contains the following columns:}
    \begin{itemize}[noitemsep]
    \item[-] {relation$\_$id (String): the unique identifier of the citation relationship between two articles;}
    \item[-] {article$\_$id$\_$citing (String): the identifier of the article which cites the other article;}
    \item[-] {article$\_$id$\_$cited (String): the identifier of the article which is cited by the other article.}
    \end{itemize}
    
    {Assumptions:}
    \begin{itemize}[noitemsep]
    \item[-] {The \textit{articles} table contains multiple entries;}
    \item[-] {The \textit{article\_author} table maps authors to articles, where one author can contribute to multiple articles, and one article can have multiple authors;}
    \item[-] {The \textit{citing\_cited} table represents citation relationships among articles in the articles table, where one article can be cited by multiple others.}
    \end{itemize}
    
    {Your task involves two steps:}
    \begin{enumerate}[noitemsep]
    \item  Understand the given SQL query in the context of the database schema described above;
    \item Convert the SQL query into a clear and natural-sounding question in everyday language, as if you were reading textual articles rather than querying a database.
    \end{enumerate}
    
    {The given SQL query:}
    \{\textcolor{red}{sql\_query}\}\\
    
    {Do not refer to relation\_id or article\_id in the natural-language question.}\\
    {You must output the SQL query and the corresponding question in the following JSON format, and do not include any extra text:
    
    \smallskip
    \{``sql": ``the given SQL query", ``question": ``the generated question"\}}
    
    \end{tcolorbox}
    \caption{Prompt template for converting SQL queries to natural language questions.}
    \label{fig:prompt_sql_to_nl}
\end{figure*}

\begin{figure*}[th!]
\small
  \begin{tcolorbox}[colback=gray!20!white,colframe=NavyBlue,title=Instruction to Convert Natural Language Questions to SQL Queries,fonttitle=\bfseries, halign title=flush center]
    {You are given a database with three tables: \textit{articles}, \textit{article-author}, and \textit{citing-cited}.}\\

    {The \textit{articles} table contains the following columns:}
    
    \begin{itemize}[noitemsep]
        \item[-] article$\_$id (String): the unique identifier of the article;
    \item[-] article$\_$title (String): the title of the article;
    \item[-]title$\_$word$\_$count (Integer): the number of words in the article's title (using spaces to determine word boundaries);
    \item[-]author$\_$count (Integer): the number of authors for the article; 
    \item[-]reference$\_$count (Integer): the number of references cited in the article.
    \end{itemize}
    
    {The \textit{article-author} table contains the following columns:}
    \begin{itemize}[noitemsep]
    \item[-]relation$\_$id (String): the unique identifier of the article-author relationship;
    \item[-]article$\_$id (String): the identifier of the associated article;
    \item[-]author$\_$name (String): the name of the author;
    \item[-] author$\_$position (Integer): the position of the author in the author list (starting from 0 for the first author).
    \end{itemize}
    
    {The \textit{citing-cited} table contains the following columns:}

    \begin{itemize}[noitemsep]
    \item[-]relation$\_$id (String): the unique identifier of the citation relationship between two articles;
    \item[-]article$\_$id$\_$citing (String): the identifier of the article which cites the other article;
    \item[-]article$\_$id$\_$cited (String): the identifier of the article which is cited by the other article.
    \end{itemize}
    
    {Assumptions:}
    \begin{itemize}[noitemsep]
    \item[-]The \textit{articles} table contains multiple entries;
    \item[-]The \textit{article\_author} table maps authors to articles, where one author can contribute to multiple articles, and one article can have multiple authors;
    \item[-]The \textit{citing\_cited} table represents citation relationships among articles in the articles table, where one article can be cited by multiple others.
    \end{itemize}
    
    Available core SQL commands:
    \begin{itemize}[noitemsep]
    
    \item[-] Aggregating: \texttt{MIN()}, \texttt{MAX()}, \texttt{COUNT()}, \texttt{SUM()}, \texttt{AVG()}, \texttt{DISTINCT}
    \item[-]Filtering: \texttt{WHERE}
    \item[-]Organizing: \texttt{ORDER BY}, \texttt{ASC}, \texttt{DESC}, \texttt{GROUP BY}
    \end{itemize}
    \medskip
    
    Available core SQL operators:
    \begin{itemize}[noitemsep]
    \item[-] Comparison: $=$, $>$, $<$, $>=$, $<=$, $<>$, \texttt{LIKE}
    \item[-] Arithmetic: $+$, $-$, $*$, $/$, \%
    \item[-] Logical: \texttt{AND}, \texttt{NOT}, \texttt{OR}, \texttt{BETWEEN}, \texttt{IN}
    \end{itemize}
    
    Your task is to:
    \begin{enumerate}[noitemsep]
    \item Understand the database schema described above and the given natural language question below;
    \item Convert the natural language question into a SQL query in the context of the database schema with the listed SQL commands and operators.
    \end{enumerate}
    
    The given natural language question:
    \{\textcolor{red}{question}\}
    \smallskip 
    
    Do not output relation\_id or article\_id in generated SQL query.\\
    Use the SQL commands and operators listed above.\\
    Make the generated SQL query aligned well with the natural language question.\\
    You must output the natural language question and the generated SQL query in the following JSON format, and do not include any extra text:\\
    \smallskip
    \{``question": ``the given question", ``sql": ``the generated SQL query"\}\\
    
    \end{tcolorbox}
    \caption{Prompt template for converting natural language questions to SQL queries.}
    \label{fig:prompt_nl_to_sql}
\end{figure*}

\begin{figure*}[!ht]
\small
  \begin{tcolorbox}[colback=gray!20!white,colframe=purple,title=Testing Instruction Using Full-text Articles,fonttitle=\bfseries, halign title=flush
    center]
    {Articles:}\\

    {\{\textcolor{red}{scientific articles}\}}\\
    
    {You are provided with multiple scientific articles above. Based on the information in these articles, answer the question provided below.}\\
    
    {If the answer consists of multiple components (e.g., author names, article titles, reference counts), separate them with commas.}\\
    {For example, if the answer includes two author names, your response should be in the format of `the-first-author-name, the-second-author-name'.}\\
    {When counting the number of words in article titles, use spaces to determine word boundaries. Words are spaced apart individually.}\\
    
    {Respond with only the final answer, with no additional explanation or formatting. If you cannot get the answer from the articles, just return `NULL'.}\\
    
    {Question:}\\
    
    \{\textcolor{red}{question}\}
    \end{tcolorbox}
    \caption{Prompt template using full-text articles as context.}
    \label{fig:prompt_generation_full_text}
\end{figure*}

\begin{figure*}[th!]
\small
  \begin{tcolorbox}[colback=gray!20!white,colframe=SeaGreen,title=Instruction Template to Generate Reasoning Traces,fonttitle=\bfseries, halign title=flush
    center]
    {Articles:}\\

    {\{\textcolor{red}{scientific articles}\}}\\
    
    {You are provided with multiple scientific articles above. Based on the information in these articles, answer the question provided below.}\\
    
    {If the answer consists of multiple components (e.g., author names, article titles, reference counts), separate them with commas.}\\
    {For example, if the answer includes two author names, your response should be in the format of `the-first-author-name, the-second-author-name'.}\\
    {When counting the number of words in article titles, use spaces to determine word boundaries. Words are spaced apart individually.}\\
    
    {Think step by step, and place your final answer within \textbackslash{boxed\{\}}. If you cannot get the answer from the articles, just return `NULL'.}\\
    
    {Question:}
    \{\textcolor{red}{question}\}
    \end{tcolorbox}
    \caption{Prompt template in the thinking mode given full-text articles or database tables as context.}
    \label{fig:prompt_generation_reasoning}
\end{figure*}

\begin{figure*}[th!]
\small
  \begin{tcolorbox}[colback=gray!20!white,colframe=purple,title=Testing Instruction Using Database Tables,fonttitle=\bfseries, halign title=flush center]
    {You are given three tables, named \textit{articles}, \textit{article-author}, and \textit{citing-cited}.}\\

    {The \textit{articles} table contains the following columns:}
    \begin{itemize}[noitemsep]
    \item[-] {article$\_$id (String): the unique identifier of the article;}
    \item[-] {article$\_$title (String): the title of the article;}
    \item[-] {title$\_$word$\_$count (Integer): the number of words in the article's title (using spaces to determine word boundaries);}
    \item[-] {author$\_$count (Integer): the number of authors for the article;}
    \item[-] {reference$\_$count (Integer): the number of references cited in the article.}
    \end{itemize}
    
    \{\textcolor{red}{the table of articles}\}\\
    
    {The \textit{article-author} table contains the following columns:}
    \begin{itemize}[noitemsep]
    \item[-] {relation$\_$id (String): the unique identifier of the article-author relationship;}
    \item[-] {article$\_$id (String): the identifier of the associated article;}
    \item[-] {author$\_$name (String): the name of the author;}
    \item[-] {author$\_$position (Integer): the position of the author in the author list (starting from 0 for the first author).}
    \end{itemize}
    
    \{\textcolor{red}{the table of article-author}\}\\
    
    {The \textit{citing-cited} table contains the following columns:}
    \begin{itemize}[noitemsep]
    \item[-] {relation$\_$id (String): the unique identifier of the citation relationship between two articles;}
    \item[-] {article$\_$id$\_$citing (String): the identifier of the article which cites the other article;}
    \item[-] {article$\_$id$\_$cited (String): the identifier of the article which is cited by the other article.}
    \end{itemize}
    
    \{\textcolor{red}{the table of citing-cited}\}\\
    
    {Based on the information in the tables above, answer the question provided below.}\\
    
    {If the answer consists of multiple components (e.g., author names, article titles, reference counts), separate them with commas.
    For example, if the answer includes two author names, your response should be in the format of `the-first-author-name, the-second-author-name'.}\\
    
    {Respond with only the final answer, with no additional explanation or formatting. 
    If you cannot get the answer from the tables, just return `NULL'.}\\
    
    {Question:}
    \{\textcolor{red}{question}\}
    \end{tcolorbox}
    \caption{Prompt template using database tables as context.}
    \label{fig:prompt_generation_meta_info}
\end{figure*}

\end{document}

%% file: math_commands.tex
\usepackage{amsmath,amsfonts,bm}

\def\eqref#1{equation~\ref{#1}}
\def\1{\bm{1}}

\DeclareMathAlphabet{\mathsfit}{\encodingdefault}{\sfdefault}{m}{sl}
\SetMathAlphabet{\mathsfit}{bold}{\encodingdefault}{\sfdefault}{bx}{n}